\documentclass[sigconf]{acmart}
\AtBeginDocument{%
  }

\copyrightyear{2025}
\acmYear{2025}
\setcopyright{cc}
\setcctype{by}
\acmConference[MM '25]{Proceedings of the 33rd ACM International Conference on Multimedia}{October 27--31, 2025}{Dublin, Ireland}
\acmBooktitle{Proceedings of the 33rd ACM International Conference on Multimedia (MM '25), October 27--31, 2025, Dublin, Ireland}
\acmDOI{10.1145/3746027.3755520}
\acmISBN{979-8-4007-2035-2/2025/10}
\newcommand{\tableCellHeight}{1.1}
\newcommand{\tabstyle}[1]{
	\setlength{\tabcolsep}{#1}
	\renewcommand{\arraystretch}{\tableCellHeight}
	\centering
	\footnotesize
}
\usepackage{color,colortbl}
\usepackage{tabulary}
\usepackage{booktabs}
\usepackage{multirow}
\usepackage{pifont}
\usepackage{rotating}
\usepackage{subcaption}
\usepackage{hyperref}

\def\textBF#1{\sbox\CBox{#1}\resizebox{\wd\CBox}{\ht\CBox}{\textbf{#1}}}

\newsavebox\CBox
\def\textBF#1{\sbox\CBox{#1}\resizebox{\wd\CBox}{\ht\CBox}{\textbf{#1}}}

\begin{document}

\title{EvoVLMA: Evolutionary Vision-Language Model Adaptation}

\author{Kun Ding}
\orcid{0000-0002-2256-8815}
\affiliation{%
	\institution{Institute of Automation, Chinese Academy of Sciences}
	\department{State Key Laboratory of Multimodal Artificial Intelligence Systems}
	\city{Beijing}
	\country{China}
}
\email{kun.ding@ia.ac.cn}

\author{Ying Wang}
\orcid{0000-0003-1385-3224}
\authornote{The corresponding author.}
\affiliation{%
	\institution{Institute of Automation, Chinese Academy of Sciences}
	\department{State Key Laboratory of Multimodal Artificial Intelligence Systems}
	\city{Beijing}
	\country{China}}
\email{ywang@nlpr.ia.ac.cn}

\author{Shiming Xiang}
\orcid{0000-0002-2089-9733}
\affiliation{%
	\institution{Institute of Automation, Chinese Academy of Sciences}
	\department{State Key Laboratory of Multimodal Artificial Intelligence Systems}
	\city{Beijing}
	\country{China}
}
\email{smxiang@nlpr.ia.ac.cn}

\renewcommand{\shortauthors}{Kun Ding, Ying Wang, Shiming Xiang}

\begin{abstract}
Pre-trained Vision-Language Models (VLMs) have been exploited in various Computer Vision tasks (e.g., few-shot recognition) via model adaptation, such as prompt tuning and adapters. However, existing adaptation methods are designed by human experts, requiring significant time cost and experience. Inspired by recent advances in Large Language Models (LLMs) based code generation, we propose an Evolutionary Vision-Language Model Adaptation (\textit{EvoVLMA}) method to automatically search training-free efficient adaptation algorithms for VLMs. We recognize feature selection and logits computation as the key functions in training-free VLM adaptation, and propose a two-stage LLM-assisted evolutionary algorithm for optimizing these parts in a sequential manner, effectively addressing the challenge posed by the expansive search space through a divide-and-conquer strategy. Besides, to enhance the stability and efficiency of searching process, we propose \textit{low-precision code conversion}, \textit{web based code execution} and \textit{process monitoring}, leading to a highly effective automatic algorithm design system. Extensive experiments demonstrate that the algorithms found by EvoVLMA can obtain promising results compared to previous manually-designed ones. More specifically, in the 8-shot image classification setting, the classical APE algorithm can be improved by 1.91 points in recognition accuracy. This research opens new possibilities for automating the optimization of adaptation algorithms of pre-trained multimodal models. \textcolor{blue}{Code is available at: \href{https://github.com/kding1225/EvoVLMA}{https://github.com/kding1225/EvoVLMA}}
\end{abstract}





\maketitle

\section{Introduction}
\label{submission}
Pre-training especially in visual-language domains has significantly propelled the advancement of Computer Vision (CV). Thanks to the vast amounts of paired image-text data, pre-trained Visual-Language Models (VLMs) can learn universal representations that exhibit strong generalization capabilities. Recently, the field of CV has shown a keen interest in transferring pre-trained models, e.g. CLIP~\cite{CLIP}, to various downstream tasks, including zero-shot and few-shot recognition~\cite{sammani2024interpreting,CoOp,TipAdapter}, open-vocabulary segmentation~\cite{ChenZQGYZXCE23}, leading to remarkable performance improvement.

\begin{figure}
	\centering
	\includegraphics[width=0.9\columnwidth]{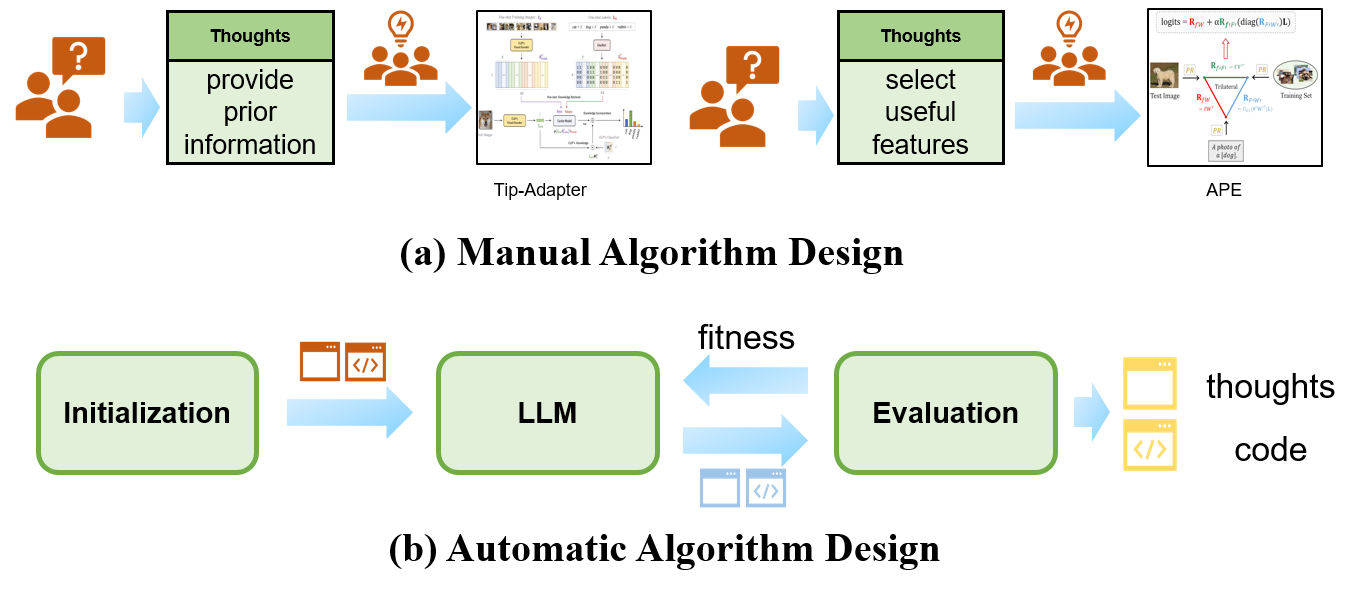}
	\caption{\textbf{Main idea of this work. }(a) The manual algorithm design. (b) The automatic algorithm design based on LLM.}
	\label{fig:main_idea}
	%	\vspace{-3pt}
\end{figure}

To adapt pre-trained VLMs to downstream tasks - particularly those with limited annotated data - parameter-efficient adaptation methods have attracted extensive research attention. Methods like prompt tuning~\cite{CoOp} optimize learnable context vectors prepended to class tokens, while adapter-based methods~\cite{TipAdapter} introduce learnable parameters in the later layers of encoders. Adapter-based methods demonstrate high computational efficiency while also achieving comparable results to prompt tuning. To further enhance the efficiency and support on-device gradient-free adaptation, training-free adaptation is drawing increased attention. For example, Zhang et al. proposed Tip-Adapter that introduces the cache model to supply prior information~\cite{TipAdapter}, Zhu et al. proposed APE that removes redundant features by the criterion of minimizing inter-class similarity~\cite{APE}, Wang et al. proposed a hard-to-beat training-free method GDA~\cite{GDA} that exploits the classical Gaussian Discriminant Analysis model. Despite their high effectiveness, these methods all depend on human expertise for algorithm design (ref. Fig.~\ref{fig:main_idea}(a)), leading to significant time costs and the requirement for specialized knowledge. Consequently, the lengthy process of algorithm development impedes the swift advancement of this field.

With the rise of LLMs like ChatGPT, researchers have gradually developed an interest in leveraging these models to accelerate the progress of scientific research. This trend has led to diverse applications of LLMs across the research lifecycle~\cite{LLM4SR}. For instance, to streamline literature reviews, Li et al. introduced an LLM-based agent system for automated literature generation~\cite{ChatCite}. Baek et al. developed ResearchAgent~\cite{ResearchAgent} to iteratively generate research hypotheses. To design research plan, LLM has been integrated with algorithm design (LLM4AD)~\cite{abs-2410-14716}. In peer review, Lu et al. proposed Agent Reviewers~\cite{luagent}, a multi-agent multimodal system for accelerating manuscript evaluation. Notable progress has also been made in combinatorial optimization, where methods like FunSearch~\cite{funsearch} and EoH~\cite{EoH} leverage LLMs to generate innovative code-based solutions for heuristic design. Building on this momentum, our work explores the integration of LLMs with algorithm design for VLM adaptation, contributing to further propel the automation trend in the formulation of research designs within scientific workflows. As illustrated in Fig.~\ref{fig:main_idea}(b), our idea is to leverage the powerful code generation capability of LLM~\cite{jiang2024survey} to automatically produce key code and continuously optimize it by feedback from code evaluation.

\begin{figure*}
	\centering
	\includegraphics[width=0.95\linewidth]{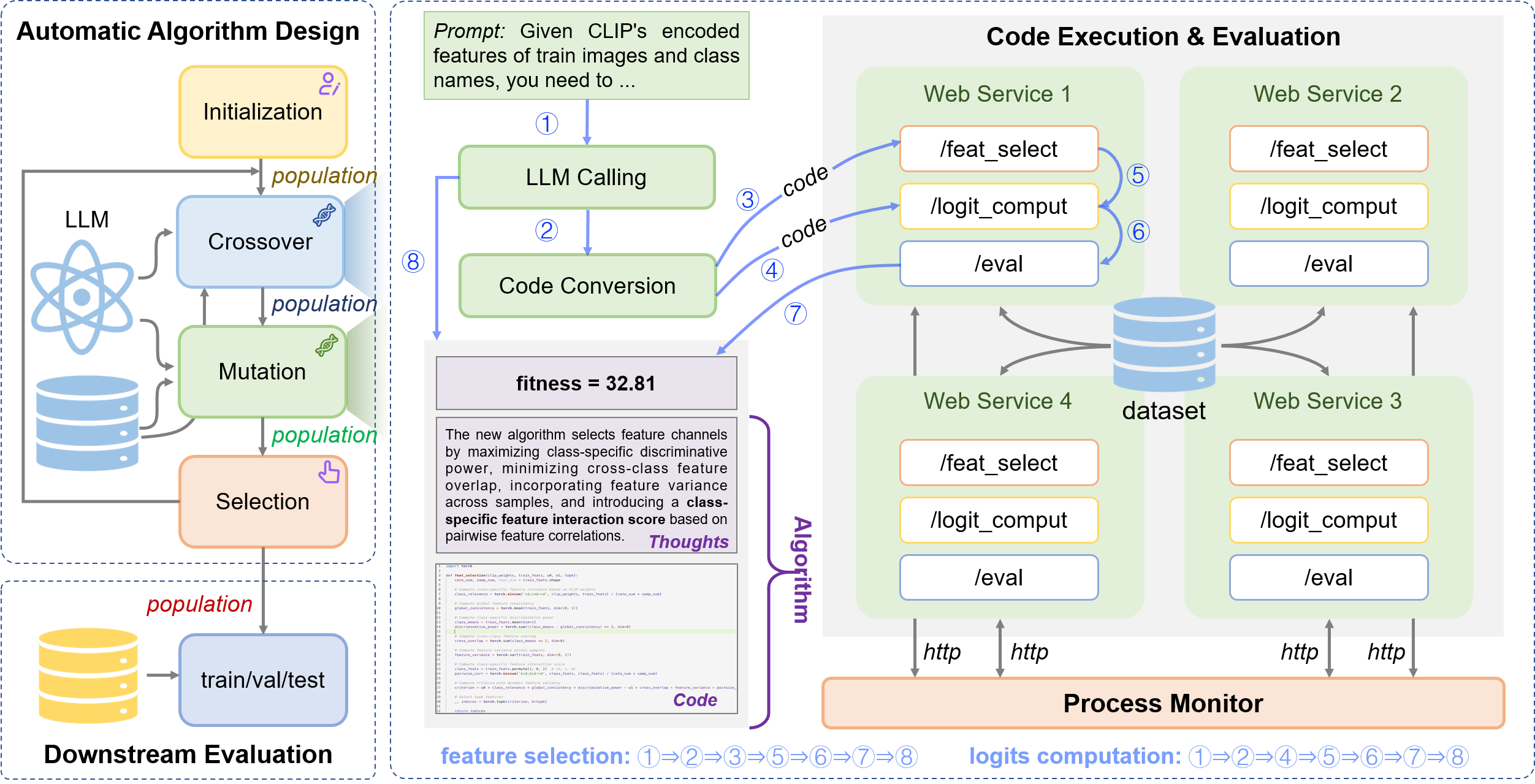}
	\caption{\textbf{The flowchart of EvoVLMA. }\textbf{Top-left panel:} the workflow of automatic algorithm designing; \textbf{right panel:} the implementation of crossover and mutation; \textbf{bottom left panel:} evaluating the performance of searched algorithms on downstream tasks. The \textbf{population} consists of several \textbf{individuals} or \textbf{algorithms}, which could be regarded as the message passed between two consecutive steps. The algorithm is composed of two components: \textbf{thoughts}, \textbf{code}.}
	\label{fig:system}
	\vspace{-2pt}
\end{figure*}

Accordingly, we propose EvoVLMA for designing efficient adaptation algorithms of pre-trained VLMs. As shown in the top-left panel of Fig.~\ref{fig:system}, its main components include: \textit{initialization}, \textit{crossover}, \textit{mutation} and \textit{selection}. The initialization step constructs an initial population consisted of existing algorithms, which serve as the starting point of searching. Crossover and mutation are specifically-designed evolutionary operators based on LLM's powerful code generation ability, which are the basic for evolving the population. To effectively select the best individuals in the population, we design a dedicated fitness function for the adaptation task. To address the distinct challenges of efficiency and stability while executing and evaluating the searched algorithm on GPU, we propose web based code execution and evaluation, while monitoring the status of execution process. We study the effectiveness of searched algorithms on several public datasets. The results demonstrate that the found algorithms have good generalization ability across different initialization, data distributions and backbones.

We summarize our contributions as follows: 1) We introduce automatic algorithm design to the domain of efficient adaptation of pre-trained VLMs and propose a two-stage evolutionary method performing algorithm searching in code space; 2) We propose web based code execution and evaluation method, and design a process monitor, boosting the efficiency and stability significantly; 3) We perform extensive experiments on several public datasets and demonstrate the efficacy of the found algorithms.

\section{Related Work}

\textbf{Vision-Language Models.} Beyond supervised pre-training~\cite{ViT,0001DMPHGSCGAJB23} and self-supervised pre-training~\cite{MAE, SimCLR, DINOv2}, multi-modal pre-training is gradually becoming a new pre-training paradigm in CV. Due to the impressive zero-shot generalization performance provided by intensive learning on massive paired image-text data crawled from internet, pre-trained Vision-Language Models (VLMs) have become new kind of foundational models applicable to various vision tasks. According to the review paper by Zhang  et al., existing VLMs have three kinds of typical architectures: two-tower VLM~\cite{CLIP,JiaYXCPPLSLD21}, two-leg VLM~\cite{CoCa,FLAVA} and one-tower VLM~\cite{TschannenMH23,JangKJKK23}. In terms of train objective, there are two kinds of learning objectives: discriminative (image contrastive loss, image-text contrastive loss, etc.) and generative (masked image modeling, masked language modeling, etc.). Given the widespread use of the CLIP model in previous work, this paper employs CLIP as the benchmark pre-trained model and investigates automated algorithm design methods. %In fact, the methodology presented in this paper holds promise for application to other pre-trained models in the future.

\textbf{Model Adaptation.} Transferring knowledge from pre-trained models to solve CV problems has become a hot topic, with methods divided into gradient-based training-required and training-free approaches. The former often preprends learnable vectors known as prompts~\cite{VPT,E2VPT,CoOp,KgCoOp,ding2024weak,DenseCLIP,ding2024multi} or inserts learnable feature adapters in the middle or latter layers~\cite{ViT-Adapter,JieDCJ24}. These parameters often account for a small percent in the overall network, making the data efficient tuning possible. Due to the gradient-based training, these methods are usually quite slow, preventing agile model development and deployment. Recently, researchers have gradually developed an interest in adaptation methods that do not require training. The key of training-free adaptation is to design a good logits computation function that effectively exploits the information in train images' features, class names' features and test images' features. The related works include: introducing cache model~\cite{TipAdapter}, removing redundant features~\cite{APE}, exploiting the classical machine learning models~\cite{GDA,GPCache}. The training-free methods do not need gradient-based learning and are usually very efficient. However, the no-training approach poses higher demands for researchers on ingenious algorithm design. This work tries to address the difficulty of automatic algorithm design for adapting VLMs by exploiting LLM's code generation ability.

\textbf{Automatic Algorithm Design.} Algorithm design for problem solving is labor-intensive as it needs in-depth domain knowledge and expert experience. The advent of LLMs has propelled the progress of automatic algorithm design remarkably, yielding a brand-new research direction, i.e., Large Language Models for Algorithm Design (LLM4AD). To design heuristics for solving combinatorial optimization problems, ReEvo~\cite{ReEvo} and EoH~\cite{EoH} integrated LLM with Evolutionary Computation, resulting in a population-based algorithm design framework. For continuous optimization, Niki van Stein and Thomas B{\"{a}}ck proposed LLaMEA~\cite{abs-2405-20132} that validates the effectiveness of LLM in generating metaphor-based optimization algorithms. Hao et al. proposed LAEA~\cite{HaoZZ24} that introduces LLM-based surrogate models to address both regression and classification problems. Besides, LLMs were also demonstrated to be effective in automatically designing Acquisition Functions for Bayesian optimization~\cite{abs-2406-04824,YaoLCZ24}. In the ML domain, LLMs have been applied to various algorithm design tasks in task planning~\cite{HuangAPM22,LinAMPB23}, reinforcement learning~\cite{ShahEOXIL23,abs-2402-16181}, neural architecture search~\cite{ChenDS23,JawaharALD24}, graph learning~\cite{abs-2403-03962}, full-pipeline AutoML~\cite{zhang2023automl, AutoML_Agent}, and so on.

Inspired by EoH, we propose EvoVLMA, a framework of automatic algorithm design for adapting VLMs to CV tasks. The proposed framework introduces an innovative two-stage decomposition strategy coupled with evolutionary algorithm-based code space exploration. This design not only overcomes current limitations in LLMs' ability to generate complex code structures but also achieves an optimal balance between exploration and exploitation in the search process.

\section{Method}
The overall workflow of EvoVLMA is displayed in Fig.~\ref{fig:system}. The core idea is treating the algorithm design problem as a code searching task. Starting from an initial population of algorithms, the algorithms are iteratively improved following an evolution framework. But different from classical evolution algorithm, the crossover and mutation operators are implemented by prompting LLMs. The LLMs generate new code of algorithm by combining and transforming one or some randomly selected code(s) from the current population. To simulate the algorithm design process of human experts (ref. Fig.~\ref{fig:main_idea}(a)), we request LLMs to also generate a natural language description of the algorithm~\cite{EoH}, i.e., the thoughts. To indicate the goodness of an algorithm, we also introduce a fitness value for each algorithm. As such, the individual or the algorithm in the population consists of two components: \textit{thoughts} and \textit{code}. 

Operationally, as shown in the top-left panel, EvoVLMA loops over a sequence of operations: crossover, mutation and selection, where crossover and mutation are implemented in the right panel by LLM calling, code conversion, code execution, evaluation and process monitoring. We have recognized that both the logits computation and the feature selection algorithm are important for model adaptation~\cite{APE}, we therefore search both of them. Considering the difficulty of joint searching, we propose to search the two algorithms sequentially, meaning we fix one as the default while searching for the optimal code for the other. The workflows in each iteration shown in the right panel for searching them are quite similar, only differing in the third step. At each iteration, the selection operator selects top-k best algorithms according to the fitness evaluated on a dataset, preventing the excessive growth of the population. After several iterations, we obtain the optimal algorithms and evaluate their performance on downstream tasks (ref. bottom-left panel). The following details every parts of EvoVLMA.
%To conduct an extensive evaluation of the searched algorithms, we use several public few-shot image recognition datasets, which are commonly adopted in previous works.

\begin{table*}[!tb]
	\caption{\textbf{Summary of initialization.} Blue color indicates extra parameters, red color indicates extra hyper-parameters. `w0', `w1', `topk', `alpha0' are hyper-parameters from the original algorithms.}
	\label{tab:init}
	\scriptsize
	\renewcommand{\arraystretch}{1.2}
	\begin{tabular}{p{2cm}p{1cm}p{10cm}p{3cm}}
		\toprule
		\textbf{Type} & \textbf{Paper} & \textbf{Thoughts} & \textbf{Function Definition} \\
		\midrule
		feature selection & APE &The feature selection algorithm defines a criterion that aims to extract the feature channels that minimize the inter-class similarity of the concatenated features of visual and category textual features, but maximize the variance of category textual features.
		& feat\_selection(clip\_weights, train\_feats, w0, w1, topk)\\
		\midrule
		logits computation & APE & The algorithm sums up two logits: the logits generated by zero-shot classifier and the logits generated by a cache model. The first logits are computed by applying linear transformation to test features. The second logits are obtained by first computing the similarity matrix between test and train features and then multiplying the transformed similarity matrix to soft train label matrix. While computing image-image similarity, selected feature channels are used.
		& compute\_logits(train\_feats, train\_labels, test\_feats, clip\_weights,
		indices, alpha0, alpha1, alpha2)\\
		\midrule
		logits computation & Tip-Adapter & The algorithm sums up two logits: the logits generated by zero-shot classifier and the logits generated by a cache model. The first logits are computed by applying linear transformation to test features. The second logits are obtained by first computing the similarity matrix between test and train features and then multiplying the transformed similarity matrix to train label matrix.
		& compute\_logits(train\_feats, train\_labels, test\_feats, clip\_weights,
		\textcolor{blue}{indices}, alpha0, \textcolor{red}{alpha1}, \textcolor{red}{alpha2})\\
		\midrule
		logits computation & GDA & The logits consist of two parts: the logits computed by CLIP's zero-shot classifier and the logits computed by Gaussian Discriminant Analysis (GDA) model. In each part, all feature channels are used. GDA is a probabilistic generative model for classification that assumes all classes are generated by Gaussian distributions with a common covariance matrix but different mean vectors. GDA first computes per-class mean vector and then estimates the inverted covariance matrix. After that the weight and bias of the GDA classifier can be computed.
		& compute\_logits(train\_feats, train\_labels, test\_feats, clip\_weights,
		\textcolor{blue}{indices}, alpha0, \textcolor{red}{alpha1}, \textcolor{red}{alpha2})\\
		\bottomrule
	\end{tabular}
\end{table*}

\subsection{Initialization}
The initialization step aims to create an initial population of algorithms. Considering that the searching space of code is quite large, directly searching the algorithm without providing any prior knowledge could generate sub-optimal solution~\cite{abs-2405-20132,ChenDS23,lu2024discovering,tnGPS}. As such, we use the public code of existing algorithms (e.g., Tip-Adapter~\cite{TipAdapter}, APE~\cite{APE}, GDA~\cite{GDA}) as the initial code. Besides, we further introduce extra parameters and hyper-parameters into the function definition. The initialization with existing methods are listed in Table~\ref{tab:init}, whose core ideas are summarized as follows: 1) For feature selection algorithm, we adopt the original code of APE as it supports feature selection. No additional parameters and hyper-parameters are introduced. 2) For logits computation algorithm, different existing algorithms can be used as initialization. While using Tip-Adapter and GDA, we add the indices of selected features `indices' as an extra parameter. Additional hyper-parameters are introduced, i.e., `alpha1' and `alpha2'. Note that not all hyper-parameters must be used in the generated code. The permission of using extra hyper-parameters makes it easier for LLMs to integrate different aspects of novel design.

An algorithm in the population should also include the thoughts indicating the underlying idea of the algorithm. For simplicity, we compose the thoughts of each algorithm by referring to the corresponding paper, which are listed in the third column of Table~\ref{tab:init}. As for the fitness value of the initial algorithm, we execute the corresponding code on holdout image recognition datasets, which are different from the datasets used for the downstream validation.

\subsection{Crossover and Mutation}
\textbf{Crossover.} This operator explores new algorithms via obtaining inspiration from multiple parent algorithms. The parent algorithms are randomly selected from the algorithms in the existing population. For this aim, the current algorithms in the population are sorted according to their fitness in ascending order. Then, two algorithms are  randomly selected with probability $p_i\propto1/(r_i+N)$, where $r_i$ is the rank of the $i$-th algorithm and $N$ is the population size. Once the parent algorithms are obtained, we instruct LLM to generate a new algorithm, which includes the thoughts and the corresponding Python code. We encourage LLM to explore novel ideas by explicitly informing LLM to generate new algorithms with different form from the parent algorithms.

The prompt template of crossover is given in appendix. The construction logic of the prompt template follows a structured approach: \textit{Firstly}, it necessitates a clear definition of the task at hand, ensuring that the objective is unequivocally communicated. \textit{Secondly}, it involves presenting example algorithms, from which LLM can draw inspiration. \textit{Thirdly}, the design requirements are elaborated in detail, specifying constraints, definition of inputs and outputs, etc. \textit{Lastly}, additional pertinent information is included to provide a comprehensive view and encourage critical thinking. By adhering to this structured format, the prompt template ensures that LLMs are well-equipped to tackle the problem, fostering both comprehension and creativity in their responses.

\textbf{Mutation.} This operator generates new algorithms by probabilistically selecting and modifying existing algorithms from the population, where each candidate is chosen with probability $p_i$. The prompt template for instructing LLM in this case is similar to that for crossover operation, which is shown in appendix. Note that we have also asked LLM to generate novel algorithm as much as possible.

In the prompt templates for both crossover and mutation, the content in braces should be filled with different contents for feature selection and logits computation. For feature selection, we ask LLM to select the best feature channels in task description, while for logits computation we ask it to devise a good function for computing classification logits. We encourage LLM to generate novel algorithm by asking it to design algorithm different from the ones in literature. The input and output information are also slightly different in these two cases. In the description of inputs and outputs, their dimensions are explicitly given, which would be helpful for LLM to generate valid expression. 

In the part of other information, we additionally tell LLM what kind of code is better. First, to avoid randomness, we ask LLM not to use random operations in the code, improving the reproducibility. Second, to avoid generating slow code, we ask LLM to avoid deep nested loops. Using learnable variables contradicts our intention to design a training-free algorithm, we also ask it to avoid this. Finally, we ask LLM to notice the readability of the generated code.

\subsection{Low-precision Code Conversion}
By prompting LLM, we can obtain the response text. Considering the unstructured characteristic of text data, we need to process the text to obtain the generated algorithm. Here, we exploit regular expression to extract the thoughts and the code part. If either the thoughts or the code cannot be parsed successfully, we invoke LLM again for another try. Besides, considering fp16 is faster than fp32, we convert the code to support fp16 computation. Here, we propose a lookup table based code conversion method. Specifically, we pre-define a lookup table containing the original functions and their converted versions. Before executing the code, we convert the code accordingly. The lookup table can be found in appendix.

\subsection{Code Execution and Evaluation}
\textbf{Evaluation.} To evaluate the quality of the generated code, we compute the fitness value by executing it with a holdout image classification dataset as an extra input. We define the classification error as  the fitness value. Let us denote the train and test set of the holdout dataset as $\mathcal{D}_{tra}$ and $\mathcal{D}_{tst}$, respectively. $\theta_0=\{w_0, w_1, k\}$ collects the hyper-parameters in feature selection code, and $\theta_1=\{\alpha_0, \alpha_1, \alpha_2\}$ collects the hyper-parameters in logits computation. Note that the specific meaning of the hyper-parameters in $\theta_0$ and $\theta_1$ could change during the searching process. The fitness value  is defined as
\begin{align}
	\text{fitness}\,=\,1\,-\,\arg\max_{\theta\in \mathcal{S}}\text{Acc}(\text{logits}(\mathcal{D}_{tst}, \mathcal{I}, \theta_1), \mathcal{D}_{tst}), \nonumber
\end{align}
where $\mathcal{I} = \text{FS}(\mathcal{D}_{tra}, \theta_0)$ denotes the indices of selected features, $\theta=\theta_0\cup\theta_1$ collects all hyper-parameters, $\mathcal{S}$ is the searching space of $\theta$. $\text{logits}(\cdot)$, $\text{Acc}(\cdot)$ and $\text{FS}(\cdot)$ are the logits computation function, the accuracy computation function and the feature selection function, respectively. We use the maximal test accuracy in searching space $\mathcal{S}$ to prevent our method from merely tuning hyper-parameters instead of generating creative algorithms. The proposed fitness equation is general to support different demands. While evaluating the feature selection function, we use the logits computation function of APE as $\text{logits}(\cdot)$. While computing classification logits by $\text{logits}(\cdot)$, the input $\mathcal{I}$ could be optional.

\textbf{Execution.} To optimize search efficiency, the code of the functions $\text{FS}(\cdot)$, $\text{logits}(\cdot)$ and $\text{Acc}(\cdot)$ are executed on GPU. However, during the search process, certain generated code may trigger CUDA errors that elude Python's `try ... exception' exception handing. These errors disrupt subsequent code execution, ultimately leading to a failure. To mitigate this issue, we encapsulate $\text{FS}(\cdot)$, $\text{logits}(\cdot)$ and $\text{Acc}(\cdot)$ as independent web services, i.e., `/feat\_select', `/logit\_comput' and `/eval', respectively (see Fig.~\ref{fig:system}). Each service operates within its own process, monitored by a dedicated process monitor. The monitor detects unresponsive processes and initiates timely restarts ensuring continuous operation. Additionally, leveraging web servers like Gunicorn, we deploy multiple instances of each service to enable parallel code execution. This architectural choice not only enhances fault tolerance but also significantly accelerates the search process through concurrent evaluations.

\textbf{Process Monitor.} As shown in Fig.~\ref{fig:system}, we invoke an individual process to monitor the worker processes of code execution. The monitoring process regularly requests the `/feat\_select', `/logit\_comput' and `/eval' services, asking them to execute test codes. Once the response fails, we kill the process of the corresponding service, immediately. Due to the features of Gunicorn, a new web service will be started, keeping the total number of worker processes fixed.

\subsection{Selection}
At each iteration of EvoVLMA, we exploit the selection operator to keep optimal individuals, avoiding the explosion in population size simultaneously. In this work, we select top-$N$ individuals with the lowest fitness value to form the new population. Here, $N$ is the population size. After this, we will enter the next iteration.

\section{Experiment}
\subsection{Settings}
To demonstrate the generalization ability of searched algorithm, we use two settings: few-shot recognition and domain generalization. The first setting compares few-shot recognition accuracy on the test set of downstream datasets. The second studies if the model fitted on source domain can be transferred to target domain.

\textbf{Datasets.} Following previous works, we use 11 publicly available datasets as the downstream datasets in the few-shot recognition setting: ImageNet~\cite{ImageNet}, Caltech101~\cite{Caltech101}, UCF101~\cite{UCF101}, EuroSAT~\cite{EuroSAT}, SUN397~\cite{SUN397}, DTD~\cite{DTD}, Food101~\cite{Food101}, OxfordPets~\cite{OxfordPets}, StanfordCars~\cite{StanfordCars}, Flowers102~\cite{Flowers102}, FGVCAircraft~\cite{FGVCAircraft}. Under the domain generalization setting, we follow previous works, where ImageNet is used as the source domain, ImageNetV2~\cite{ImageNetV2} and ImageNet-Sketch~\cite{ImageNetS} are used as the target domains. For algorithm design, we adopt CIFAR100~\cite{krizhevsky2009learning}, FashionMnist~\cite{FashionMnist}, ObjectNet~\cite{ObjectNet}, UcMerced\footnote{http://weegee.vision.ucmerced.edu/datasets/landuse.html} and UCSDBirds\footnote{https://www.vision.caltech.edu/datasets/cub\_200\_2011} as holdout datasets.

\textbf{Compared Methods.} As we target at designing training-free adaptation methods, we mainly compare the searched algorithms with existing algorithms designed by human experts, including Tip-Adapter~\cite{TipAdapter}, APE~\cite{APE} and GDA~\cite{GDA}. It is worth noting that GDA is currently the SOTA method, renowned for its unbeatable performance.

\textbf{Implementation Details.} For automatic algorithm design, we use multiple holdout datasets. Each holdout dataset is split to a train set and a test set. We randomly sample 1, 2, 4, 8, 16 samples as the few-shot train sets. We execute the code on the few-shot train sets and evaluate the accuracy on the test sets. The final accuracy is averaged on different holdout datasets and different shots. The searching space $\mathcal{S}$ is defined as the Cartesian product of searching space of each hyper-parameter, i.e., $\{0.5\}^2\times \{0.7d\}\times \{0.1, 1, 10\}^3$, where $d$ denotes the feature dimension of CLIP's image features. In the proposed method, we adopt a two-stage approach to search for feature selection algorithm and logits computation algorithm, separately. We evaluate the impact of the order in which they are applied on the results. In each stage, the population size and number of iterations are set to 10. For code generation, we adopt DeepSeek~\cite{deepseek} as the LLM due to its affordable price.

For downstream evaluation, we split each dataset to train, validation and test set following APE~\cite{APE}. For each searched or manual algorithm, we search the optimal hyper-parameters on the validation set using Optuna~\cite{optuna}. Without further specification, we use the TPESampler sampler and 500 trials. We find that Optuna based hyper-parameter optimization can achieve comparable performance to grid searching but with lower time cost.

Without further specification, we use ResNet-50 as the image encoder. We also evaluate the effectiveness using different backbones. For performance evaluation, we report the top-1 accuracy on the test set of the downstream datasets. We report results with different shots (1-shot, 2-shot, 4-shot, 8-shot and 16-shot) to study the effectiveness using different numbers of train samples. To reduce randomness, the average results over three seeds are reported.

\begin{figure*}[thb]
	\centering
	\includegraphics[width=0.95\linewidth]{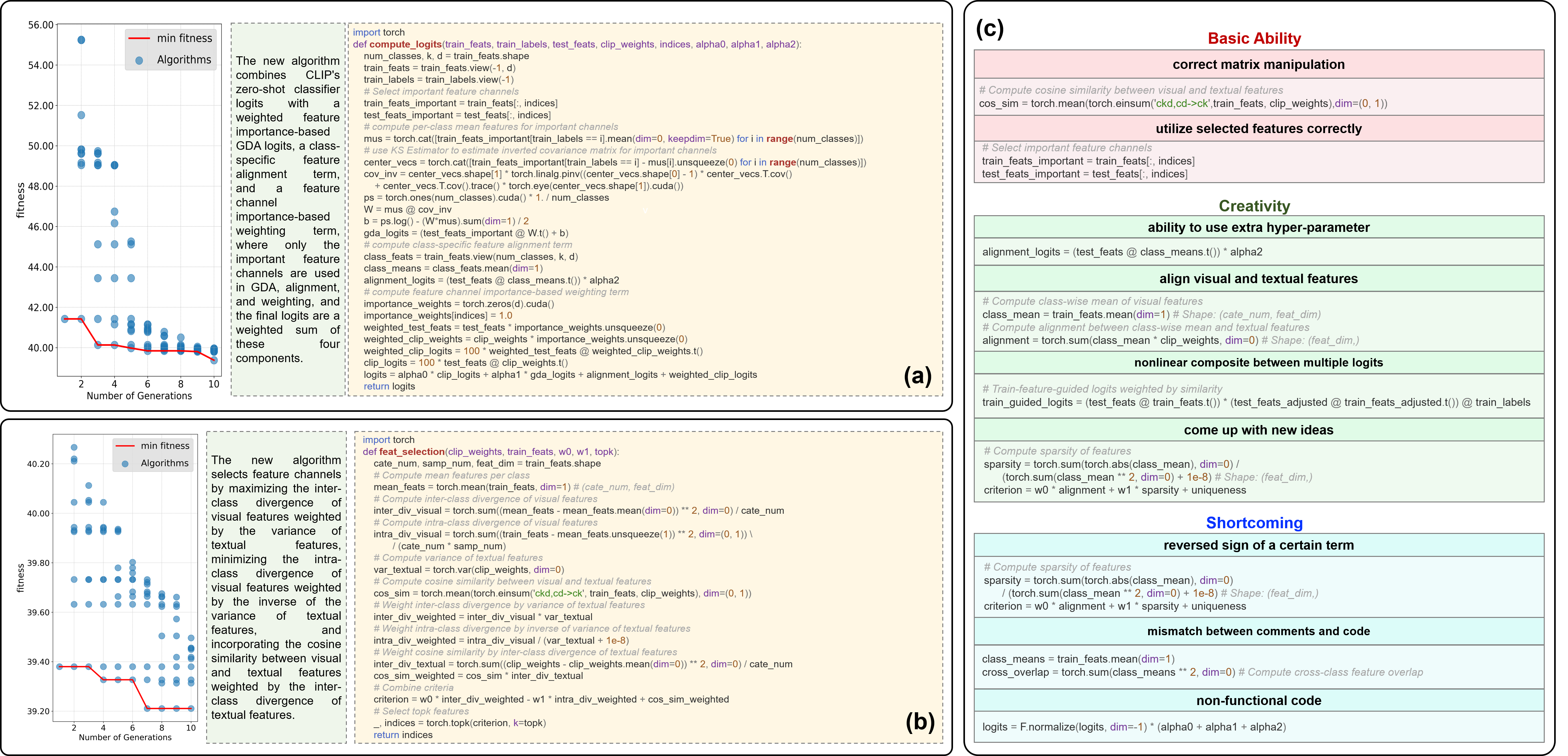}
	\caption{Searching process, found algorithms and code snippets. (a) Object curve and found logits computation algorithm using GDA based initialization (stage 1 of `logit$\rightarrow$fs'). (b) Object curve and found feature selection algorithm using GDA based initialization (stage 2 of `logit$\rightarrow$fs'). (c) Examples of searched code snippets.}
	\Description{Searching process, found algorithms and code snippets.}
	\label{fig:GDA_5di1nt10_logit+fs_vis}
\end{figure*}

\begin{table*}[thb]
	\tabstyle{10pt}
	\caption{\textbf{Few-shot classification and domain generalization results.} TIP: Tip-Adapter, IN: ImageNet, V2: ImageNet-V2, S: ImageNet-S. The bold numbers represent the accuracy gap between manually-designed and automatically-generated code.}
	\label{tab:few_shot}
	\renewcommand{\arraystretch}{1.15}
	\begin{tabular}{ccccccccccccc}
		\toprule
		\multirow{2}{*}{\textbf{Method/Init}}&\multirow{2}{*}{\textbf{Strategy}}&\multicolumn{6}{c}{\textbf{Few-shot Classification}}&\multicolumn{4}{c}{\textbf{Domain Generalization (16-shot)}}\\
		\cmidrule(r){3-8}\cmidrule(l){9-12}
		& & 1-shot & 2-shot & 4-shot & 8-shot & 16-shot & Average & IN & V2  & S & Average\\
		\midrule
		TIP & --&62.49&	64.25&	66.71&	68.98	&70.76	&66.64& 62.75&	55.35&	36.10& 51.40\\
		TIP & fs$\rightarrow$logit & 62.72	&64.93&	67.53&	69.69&	71.17&	67.21& 62.82&	55.39&	35.97&51.39\\
		&&\textBF{+0.23} & \textBF{+0.68} & \textBF{+0.82} & \textBF{+0.71} & \textBF{+0.41}&\textBF{+0.57}&\textBF{+0.07} & \textBF{+0.04} & \textBF{-0.13} & \textBF{-0.01}\\
		TIP & logit$\rightarrow$fs &62.91	&64.76&	67.25&	69.33&	71.00&	67.05& 62.93&	55.23&	36.35	&51.50\\
		&&\textBF{+0.42} & \textBF{+0.51} & \textBF{+0.54} & \textBF{+0.35} & \textBF{+0.24} &\textBF{+0.41}&\textBF{+0.18} & \textBF{-0.12} & \textBF{+0.25} & \textBF{+0.10}\\
		\midrule
		APE & --& 63.94	&65.16&	68.23	&70.33&	72.28&	67.99& 62.92&	55.26&	36.41	& 51.53\\
		APE & fs$\rightarrow$logit &64.18	&66.23&	69.60&	72.09&	74.11&	69.24& 64.49&	56.48&	36.36 &52.44\\
		&&\textBF{+0.24}&\textBF{+1.07}&\textBF{+1.37}&\textBF{+1.76}&\textBF{+1.83}&\textBF{+1.25}&\textBF{+1.57}&\textBF{+1.22}&\textBF{-0.05}&\textBF{+0.91}\\
		APE & logit$\rightarrow$fs & 63.80&	66.07&	69.51&	72.24&	74.00&	69.12& 64.45&	56.34&	35.66	& 52.15\\
		&&\textBF{-0.14}&\textBF{+0.91}&\textBF{+1.28}&\textBF{+1.91}&\textBF{+1.72}&\textBF{+1.13}&\textBF{+1.53}&\textBF{+1.08}&\textBF{-0.75}&\textBF{+0.62}\\
		\midrule
		GDA & --& 62.55	&66.00&	70.22&	73.72&	76.45&	69.79& 62.74&	55.48&	35.67&51.30\\
		GDA & fs$\rightarrow$logit & 63.56	&66.51&	70.35	&73.99	&76.43	&70.17& 64.80&	56.45	&35.25	&52.17\\
		&&\textBF{+1.01}&\textBF{+0.51}&\textBF{+0.13}&\textBF{+0.27}&\textBF{-0.02}&\textBF{+0.38}&\textBF{+2.06}&\textBF{+0.97}&\textBF{-0.42}&\textBF{+0.87}\\
		GDA & logit$\rightarrow$fs & 62.99&	66.20&	70.02&	73.59&	76.33&	69.83&64.72	&56.22&	34.82&51.92\\
		&&\textBF{+0.44}&\textBF{+0.20}&\textBF{-0.20}&\textBF{-0.13}&\textBF{-0.12}&\textBF{+0.04}&\textBF{+1.98}&\textBF{+0.74}&\textBF{-0.85}&\textBF{+0.62}\\
		\bottomrule
	\end{tabular}
\end{table*}

\subsection{Main Results}
\textbf{Qualitative Analysis.} The searching process and the found algorithms are visualized in Fig.~\ref{fig:GDA_5di1nt10_logit+fs_vis}(a) and (b). The first column shows the searching process. No matter for searching logits computation or feature selection algorithm, the population has the trend of evolving towards to the status with lower fitness. The minimal fitness curves also reflect this point. The second column shows the thoughts and the third column shows the code. For logits computation, the new algorithm introduces important feature channels to the original `gda\_logits' term. An alignment term that multiplies per-class mean features to test features is introduced. Besides, a new weighted logits introducing important features is designed. As can be seen, LLM has the ability to exploit the provided information of important features. As for feature selection, the searched algorithm introduces weighted inter-class divergence and intra-class divergence. By this way, the selected features should be good for visual and text modalities, simultaneously. The algorithm also introduces a term `cos\_sim' to promote the alignment between visual and textual modalities, which cannot not be found in existing literature.

We further give more generated distinctive code snippets in Fig.~\ref{fig:GDA_5di1nt10_logit+fs_vis}(c), from which we find: 1) The current LLMs already have the basic abilities, such as matrix manipulation. 2) The current LLMs are capable of generating creative code snippets. However, some may not be practical, e.g., the sparsity. 3) LLMs can also generate some redundant codes as well as wrong codes that cannot implement the thoughts. The results imply that LLM-based code evolution is a promising technique for automatic algorithm design. However,  there is still a space for further improving LLM's code generation ability and enhancing the usage of this ability. 

\textbf{Quantitative Analysis.} The results of manually-designed algorithms and automatically-designed algorithms under the few-shot setting are listed in Table~\ref{tab:few_shot}. The bolded numbers highlight the improvements over manual algorithms, which imply that the automatically-designed algorithms outperform manual ones in most cases. The improvement can be as large as 1.91\%. Although GDA has already achieved SOTA results, the automatically-designed algorithms can still achieve better results. It has been recognized that GDA cannot achieve good results under low-shot settings due to the  inaccurate covariance estimation. The automatically-designed algorithms improve this notably. The optimization order of feature selection and logits computation does have effect on downstream performance. The order `fs$\rightarrow$logit' works slightly better in most cases. However, this needs more extensive testing.

The results under the domain generalization setting are listed in Table~\ref{tab:few_shot}, where the performance gaps between manual algorithms and the corresponding automatic algorithms are highlighted. As can be seen, automatic algorithms outperform manual ones on source domain, while can still achieve competitive performance on target domains in most cases. In terms of average performance, automatic algorithms can acquire better trade-off between source and target performance.

\begin{table}[t]
	\tabstyle{10pt}
	\caption{Comparison to iterative refinement.}
	\label{tab:iterative_refine}
	\renewcommand{\arraystretch}{1}
	\begin{tabular}{lccc}
		\toprule
		\textbf{iteration}& \textbf{TIP} & \textbf{APE} & \textbf{GDA}\\
		\midrule
		\multicolumn{3}{l}{\textbf{population-based (evolutionary algorithm): }}\\
		accuracy & 67.21 & 69.24 & 70.17 \\
		total token/K & 633 & 676 & 812\\
		\midrule
		\multicolumn{3}{l}{\textbf{individual-based (iterative refinement):} }\\
		accuracy & 66.77 & 66.41 & 69.22 \\
		total token/K & 701 & 1005 & 884 \\
		\bottomrule
	\end{tabular}
\end{table}

\begin{table}[t]
	\tabstyle{14.2pt}
	\caption{Two-stage searching v.s. joint searching v.s. optimizing the logits computation algorithm only.}
	\label{tab:joint_search}
	\renewcommand{\arraystretch}{0.9}
	\begin{tabular}{lccc}
		\toprule
		\textbf{Strategy}& \textbf{TIP} & \textbf{APE} & \textbf{GDA} \\
		\midrule
		baseline & 66.64 & 67.99 & 69.79 \\
		fs$\rightarrow$logit  & 67.21 & 69.24 & 70.17\\
		joint & 66.58 & 68.46 & 69.43\\
		logit & 67.21 & 68.55 & 69.84\\
		\bottomrule
	\end{tabular}
	\vspace{-5pt}
\end{table}

\subsection{Ablation Study}
\textbf{Comparison to Iterative Refinement.} This work employs a population based evolutionary algorithm for algorithm search, which distinguishes itself from the iterative refinement strategy widely adopted in existing studies such as~\cite{yang2023large}. Recent research like~\cite{liu2024large} has started to explore population-based methods, revealing that such approaches hold a distinct advantage in balancing exploration and exploitation. To further validate the effectiveness of population-based methods in our specific scenario, we implemented the iterative refinement method used in~\cite{yang2023large} to generate VLM adaptation code. Specifically, 8 in-context learning code examples are utilized, and the `fs$\rightarrow$logit' searching strategy is employed with 100 iterations allocated to each stage. The results shown in Table~\ref{tab:iterative_refine} demonstrate that the population-based method can outperform the individual-based iterative refinement approach even when using fewer tokens, which underscores the greater effectiveness of the population-based method.

\textbf{Effectiveness of Feature Selection.} In this work, we optimize both the feature selection and logits computation algorithms. Here, we compare it with the strategy that only optimizes the logits computation algorithm. With APE based initialization, the logits computation uses the indices of selected feature channels by its original feature selection algorithm. With GDA or Tip-Adapter based initialization, no feature selection is involved. The results are presented in Table~\ref{tab:joint_search}. We can see that only optimizing logits computation algorithm acquires worse results than optimizing both, implying the importance of selecting good features.

\textbf{Joint Searching.} In this work, we propose a two-stage searching strategy. Here, we validate its rationality by comparing it with joint searching strategy. In this joint strategy, we create an initial algorithm by combing the feature selection algorithm in APE and the logits computation algorithm in one of Tip-Adapter, APE and GDA. We create the code accordingly implementing this algorithm. The initial algorithm with the code is given in in appendix. Besides, in our method, we have also modified the prompt template accordingly. The results of joint searching strategy are listed in Table~\ref{tab:joint_search}, which are worse than two-stage strategy and even the logit-only searching, implying that joint searching cannot enable us to find better algorithms. Through iterative analysis of population dynamics (see appendix), we observe that contemporary LLMs struggle to generate functionally improved code when required to integrate multiple interdependent components. In conclusion, the proposed two-stage strategy cuts down the complexity of the problem effectively.

\begin{table}[t]
	\tabstyle{11.5pt}
	\caption{\textbf{Effectiveness of initializing algorithm. }`no init' denotes no initialization is used.}
	\label{tab:effect_init}
	\renewcommand{\arraystretch}{1.15}
	\begin{tabular}{ccccc}
		\toprule
		\textbf{Strategy}& \textbf{TIP} & \textbf{APE} & \textbf{GDA} & \textbf{no init}\\
		\midrule
		baseline & 66.64 & 67.99 & 69.79 & --\\
		fs$\rightarrow$logit  & 67.21 & 69.24 & 70.17 & 65.48\\
		logit$\rightarrow$fs & 67.05 & 69.12 & 69.83 & 66.18\\
		\bottomrule
	\end{tabular}
\end{table}

\begin{table}[t]
	\tabstyle{6pt}
	\caption{\textbf{Effect of holdout datasets.} C: CIFAR100, F: FashionMnist, O: ObjectNet, M: UcMerced, B: UCSDBirds}
	\label{tab:holdout_dataset}
	\renewcommand{\arraystretch}{1.16}
	\begin{tabular}{lcccccc}
		\toprule
		\multirow{2}{*}{\textbf{Datasets}}&\multicolumn{3}{c}{\textbf{fs}$\rightarrow$\textbf{logit}} & \multicolumn{3}{c}{\textbf{logit}$\rightarrow$\textbf{fs}}\\
		\cmidrule(r){2-4}\cmidrule(l){5-7}
		& \textbf{TIP} & \textbf{APE} & \textbf{GDA} & \textbf{TIP} & \textbf{APE} & \textbf{GDA}\\
		\midrule
		C+F & \textBF{67.71} & {68.39} & 69.98 & \textBF{67.24} & 68.60 & \textBF{70.05}\\
		C+F+O+M+B & {67.21} & \textBF{69.24} & \textBF{70.17}& 67.05 & \textBF{69.12} & 69.83\\
		\bottomrule
	\end{tabular}
\end{table}

\begin{table*}[t]
	\tabstyle{12pt}
	\caption{Per-shot results of few-shot recognition using different LLMs. The searching strategy is `fs$\rightarrow$logit'.}
	\label{tab:diff_LLM}
	\renewcommand{\arraystretch}{1}
	\begin{tabular}{ccccccccc}
		\toprule
		\textbf{Initialization}& \textbf{LLM} &\textbf{Error Rate}& \textbf{1-shot} & \textbf{2-shot}& \textbf{4-shot} & \textbf{8-shot} & \textbf{16-shot} & \textbf{Average}\\
		\toprule
		\multirow{5}{*}{Tip-Adapter} & -- & --& 62.49&	64.25&	66.71&	68.98&	70.76&	66.64\\
		& DeepSeek & 21.23\%& 62.72	&64.93	&67.53	&69.69&	71.17&	67.21\\
		& Qwen &39.64\%& 63.50&	65.23	&68.44&	70.92&	72.67&	68.15\\
		& GPT-3.5 &48.42\%&63.92&	65.68&	68.28&	70.96&	72.90&	68.35\\
		& GPT4o-mini &52.74\%&63.82&	65.8	&69.15	&71.88	&73.79&	68.89\\
		\midrule
		\multirow{5}{*}{APE} & -- && 63.94&	65.16&	68.23&	70.33&	72.28&	67.99\\
		& DeepSeek &27.21\%& 64.18&	66.23&	69.60&	72.09&	74.11&	69.24\\
		& Qwen &51.12\%& 63.47&	64.97&	68.38&	70.76&	72.52&	68.02\\
		& GPT-3.5 &74.96\%&63.68&	65.61&	68.50&	70.79&	72.72&	68.26\\
		& GPT4o-mini &84.91\% & 63.55&	64.85&	67.43&	70.38&	71.98&	67.64\\
		\midrule
		\multirow{5}{*}{GDA} && -- & 62.55&	66.00&	70.22&	73.72&	76.45&	69.79\\
		& DeepSeek &17.74\%& 63.56&	66.51&	70.35&	73.99&	76.43&	70.17\\
		& Qwen &54.48\%& 62.47&	65.69&	69.60&	72.61&	75.64&	69.20\\
		& GPT-3.5 &69.23\%&62.83&	65.73&	69.12&	73.03&	76.04&	69.35\\
		& GPT4o-mini &51.08\% & 63.22&	65.79	&69.56&	72.19&	73.91	&68.93\\
		\bottomrule
	\end{tabular}
	%\vspace{-2pt}
\end{table*}

\textbf{Different LLMs.} We further evaluate the effectiveness of different LLMs in the proposed method. The compared LLMs include: DeepSeek, qwen-coder-plus-2024-11-06 (denoted as Qwen), gpt-3.5-turbo (denoted as GPT-3.5), and gpt-4o-mini (denoted as GPT4o-mini). We use the `fs$\rightarrow$logit' searching strategy and the ResNet-50 backbone. The results using different existing methods (Tip-Adapter, APE and GDA) as initialization are listed in Table~\ref{tab:diff_LLM}. With Tip-Adapter based initialization, the searched algorithms with all compared LLMs surpass the manually designed algorithm, i.e., Tip-Adapter. In this case, Qwen, GPT-3.5 and GPT4o-mini work better than DeepSeek. By contrast, with APE and GDA based initialization, codes found by DeepSeek are better than those found by the rest LLMs. We also count the code execution error rate while searching codes, which is defined as the percent of wrongly executed codes in the codes being evaluated. As can be seen, DeepSeek has lower error rate while different initialization is used, implying that DeepSeek has strong ability in generating executable codes. In the future, it is worth conducting a more extensive evaluation of existing LLMs using more diverse datasets.

\begin{figure}[t]
	\centering
	\begin{subfigure}{.48\columnwidth}
		\centering
		\includegraphics[height=3.4cm]{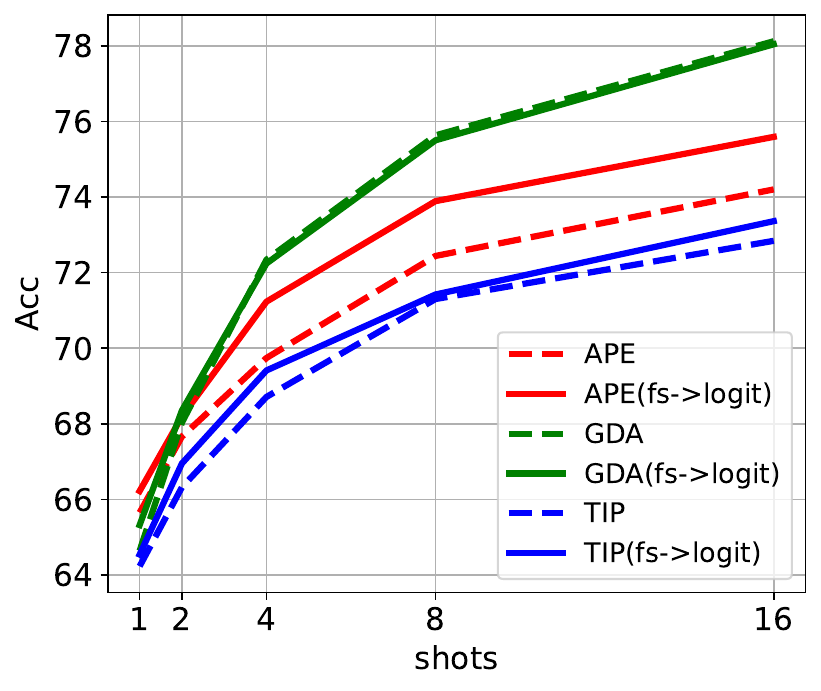}
	\end{subfigure}%
	~
	\begin{subfigure}{.48\columnwidth}
		\centering
		\includegraphics[height=3.4cm]{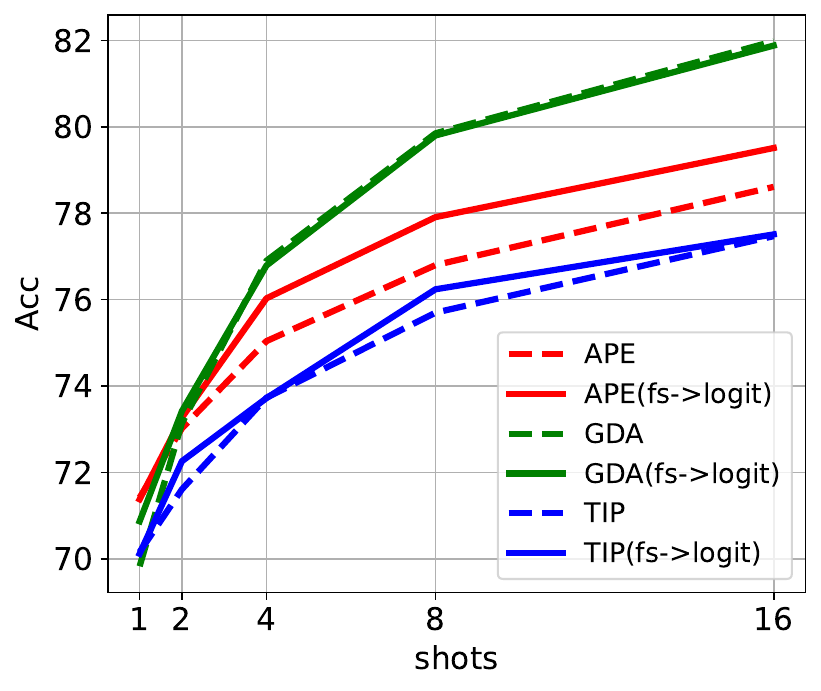}
	\end{subfigure}%
	\caption{\textbf{Effect of different visual backbones.} Left to right: ResNet-101 and ViT-B/16.}
	\label{fig:bb}
	\vspace{-5pt}
\end{figure}

\begin{table}[t]
	\tabstyle{16pt}
	\caption{Effect of iteration number.}
	\label{tab:alter_search}
	\renewcommand{\arraystretch}{1.17}
	\begin{tabular}{ccccc}
		\toprule
		\textbf{iteration}& \textbf{TIP} & \textbf{APE} & \textbf{GDA}\\
		\midrule
		1 & 67.21 & 69.24 & 70.17\\
		2 & 67.65 & 69.31 & 70.37\\
		3 & 67.77 & 69.43 & 70.39\\
		\bottomrule
	\end{tabular}
    \vspace{-7pt}
\end{table}

\textbf{Initialization.} To demonstrate the effectiveness of algorithm initialization, we compare the average few-shot classification accuracy across 1, 2, 4, 8, 16 shots with and without initialization in Table~\ref{tab:effect_init}. Clearly, while not initializing the algorithm, notable degeneration in accuracy w.r.t. the worst Tip-Adapter based initialization occurs. This suggests that current LLMs still has limited capabilities in designing algorithms from the ground up. Introducing more domain knowledge to LLMs could be a promising solution.

\textbf{Datasets.} The effect of holdout datasets are given in Table~\ref{tab:holdout_dataset}. The average few-shot classification accuracy over 1, 2, 4, 8, 16 shots is reported. Using 2 datasets and 5 datasets can obtain similar downstream performance, which implies that code evolution should be not very sensitive to dataset scale.

\textbf{Backbone.} The results with different vision backbones are shown in Fig.~\ref{fig:bb}. As can be seen, improved accuracies can also be observed while transferring the algorithms found on ResNet-50 feature distribution to ResNet-101 and ViT-B/16 feature distributions. This implies that the searched algorithms have good robustness to visual feature distribution drifting.

\textbf{Alternating Optimization Strategy.} Our framework alternately optimizes feature selection and logits computation in an iterative manner. To investigate the impact of iteration numbers, we conduct experiments with multiple optimization cycles, with results presented in Table~\ref{tab:alter_search}. The empirical findings reveal that while additional iterations yield marginal performance gains, the improvement diminishes significantly after the first iteration. Therefore, to maintain computational efficiency while ensuring competitive performance, we adopt a single-iteration configuration for all experiments. This design choice effectively balances model performance with experimental expediency.

\section{Conclusion}
This work proposes an LLM-based two-stage code evolution method EvoVLMA to address the automatic algorithm designing problem for VLM adaptation. The core ingredients of this method include crossover, mutation and selection, all of which are implemented by prompting LLM. EvoVLMA is flexible to support different searching strategies, including two-stage and joint searching. We have validated the effectiveness of searched algorithms on several downstream datasets with different initialization, strategies, and settings.

As a preliminary exploration, we can conclude that it is possible to utilize LLMs to design model adaptation algorithms automatically. Other areas in CV that require manual algorithm design may also be inspired by our work. However, we must also soberly recognize the limitations of current LLMs in automatic algorithm design, including their dependence on good initialization, constrained creativity, and challenges in generating complex multi-function code. Addressing these issues will be the direction for future work.

\begin{acks}
This work was supported by the Strategic Priority Research Program of Chinese Academy of Sciences (Grant No. XDA0480200), the National Natural Science Foundations of China (Grant No. 62306310) and the Beijing Natural Science Foundation (Grant No. L242093).
\end{acks}

\bibliographystyle{ACM-Reference-Format}
\balance
\bibliography{example_paper}

\clearpage
\appendix
\setcounter{section}{0}
\renewcommand{\thesection}{A}

\section*{Appendix}

\subsection{Low-precision Code Conversion}
To enable fp16 inference, we post-process the searched code by modifying some PyTorch function callings. For the functions listed in the first column of Table~\ref{tab:code_convert}, we replace them with the corresponding functions in the right column. For example, the original `torch.linalg.pinv' does not support fp16 computation, we define a new function calling it after converting the data type to fp32. The results are converted back to fp16 format. In the downstream stage, we can also use such conversion for speedup.

\subsection{Hyper-parameter Optimization}
For evaluating the searched code on downstream datasets, we have adopted Optuna for hyper-parameter optimization. The searched hyper-parameters and the searching ranges are listed in Table~\ref{tab:hyper_params}. In this table, $d$ denotes the number of feature dimensions, $topk$ denotes the number of kept feature channels. The searching strategies `fs$\rightarrow$logit', `logit$\rightarrow$fs', `joint' and `logit' denote searching feature selection algorithm first and then logits computation algorithm, searching logits computation algorithm first and then feature selection algorithm, searching them jointly and only searching the logits computation algorithm, respectively. By default, the number of trials is set to 500. However, it will be slow for GDA and the searched algorithms using GDA as initialization, we set the number of trials as 100 in this cases.

\subsection{Initialization}
\begin{figure*}[t]
	\centering
	\includegraphics[width=0.95\linewidth]{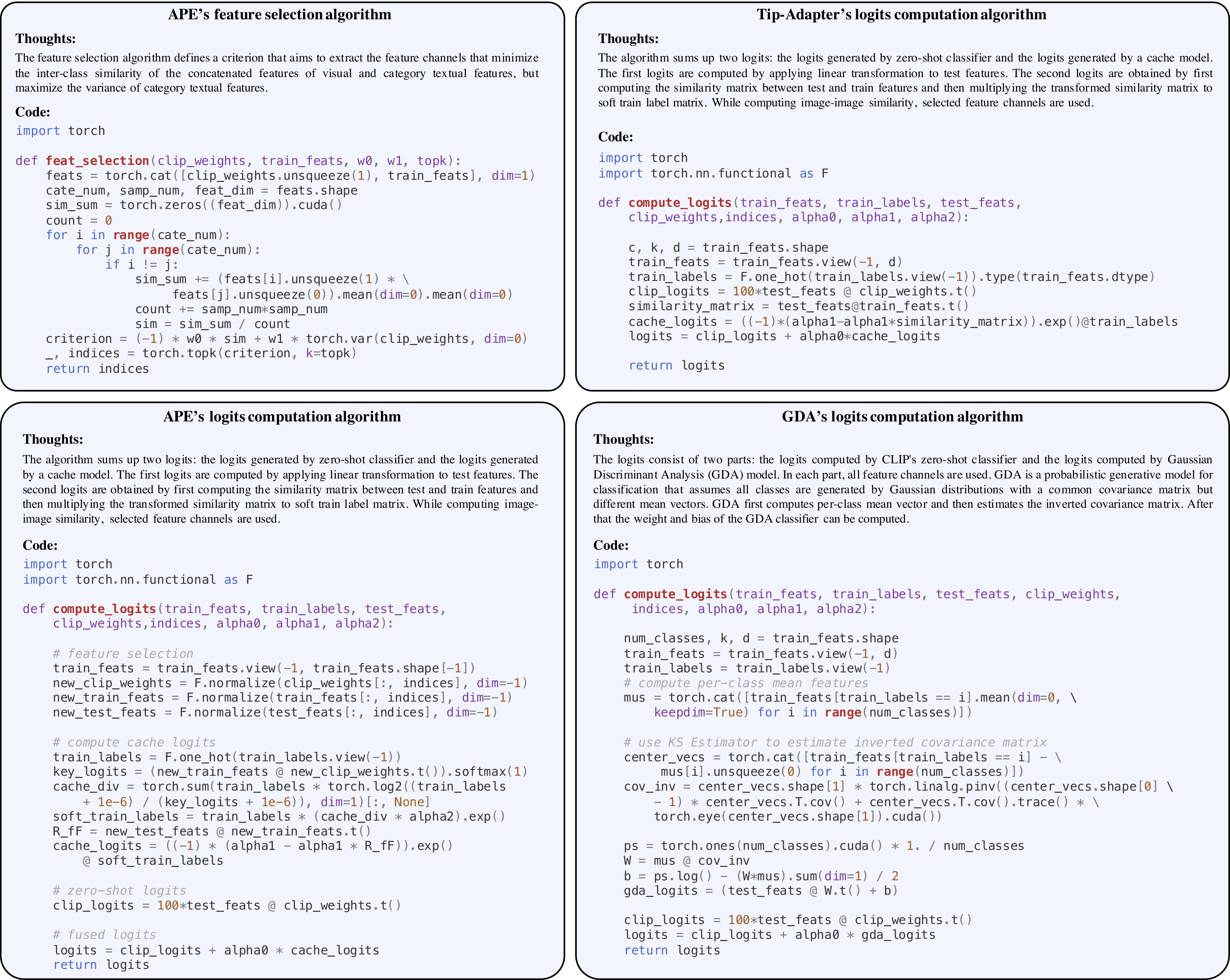}
	\caption{Initial algorithms for two-stage searching. }
	\label{fig:init_algo}
	\vspace{-5pt}
\end{figure*}

The initial algorithms for two-stage searching are shown in Fig.~\ref{fig:init_algo}. We additionally construct the description of thoughts for each initial algorithm. While initializing the feature selection algorithm using APE, no extra parameters and hyper-parameters are introduced. While initializing the logits computation algorithm using APE, we additionally introduce `alpha1' and `alpha2' as hyper-parameters. While initializing the logits computation algorithm using Tip-Adapter and GDA, we additionally introduce `indices' as parameter, `alpha1' and `alpha2' as hyper-parameters. `indices' denotes the indices of selected feature channels.

\begin{figure*}[t]
	\centering
	\includegraphics[width=0.95\linewidth]{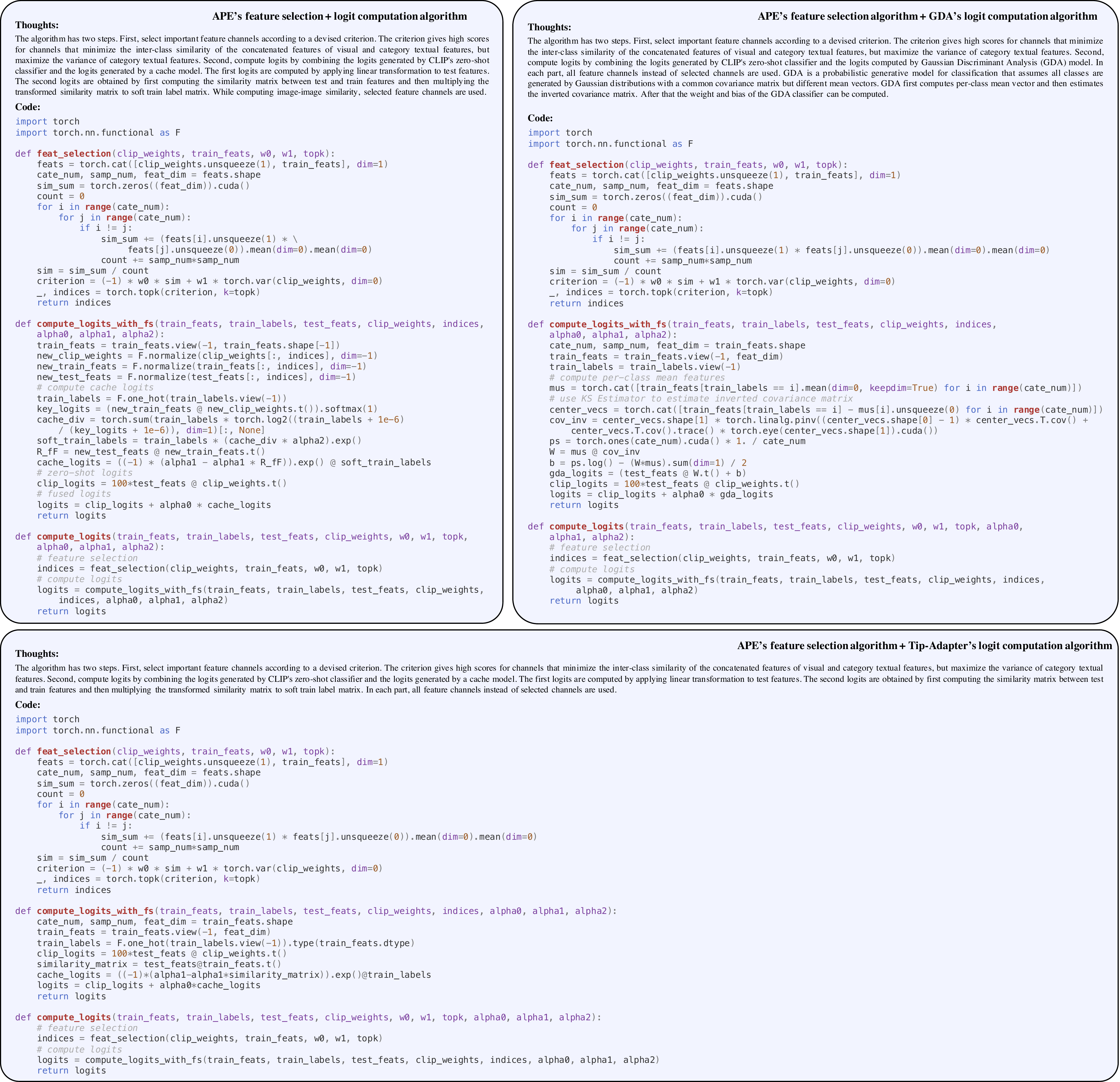}
	\caption{Initial algorithms for joint searching. }
	\label{fig:init_joint}
	\vspace{-8pt}
\end{figure*}

The initial algorithms for joint searching are shown in Fig.~\ref{fig:init_joint}. To enable simultaneous searching of feature selection and logits computation algorithms, we construct an overall algorithm integrating them together. In the description of thoughts, we indicate that the algorithm consists of two steps: 1) selecting important features; 2) computing logits. The corresponding code includes three functions: `feat\_selection', `compute\_logits\_with\_fs', and `compute\_logits'. The `feat\_selection' function selects features based on CLIP's text features `clip\_weights' and train images' features `train\_feats'. It returns the indices of selected feature channels `indices'. The `compute\_logits\_with\_fs' function computes classification logits of test features `test\_feats' by exploiting the train images' features `train\_feats', the indices of selected feature channels `indices', and so on. The `compute\_logits' calls the above two functions to compute classification logits. Note that, the `compute\_logits\_with\_fs' is constructed based on the original code of each algorithm. As Tip-Adapter and GDA do not use selected features, the `indices' parameter is not used in the initial code.

\subsection{Prompts}
\begin{figure*}[t]
	\centering
	\includegraphics[width=0.95\linewidth]{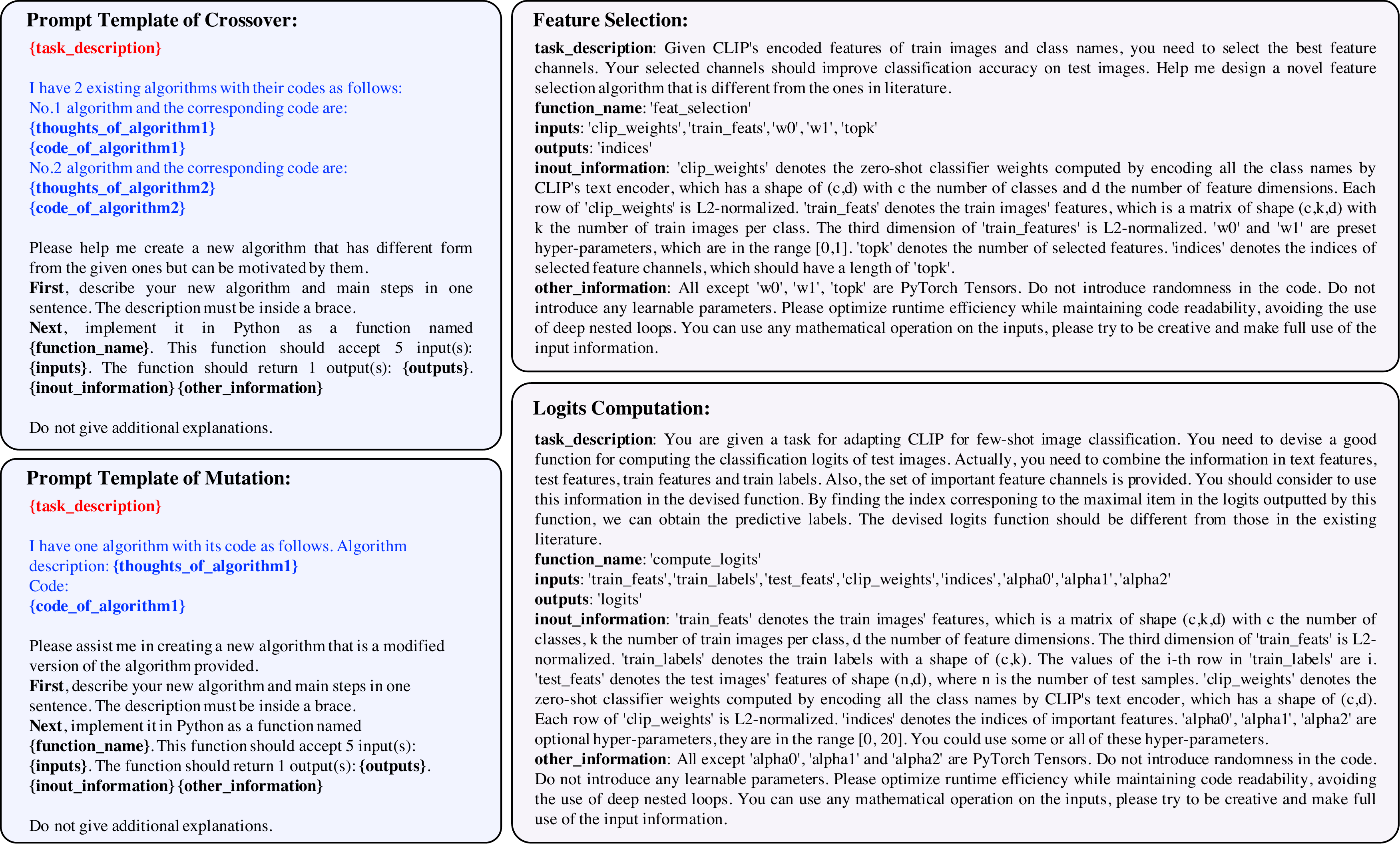}
	\caption{Prompts of crossover and mutation for searching feature selection and logits computation algorithms.}
	\label{fig:prompt}
	\vspace{-5pt}
\end{figure*}

\begin{figure*}[t]
	\centering
	\includegraphics[width=0.95\linewidth]{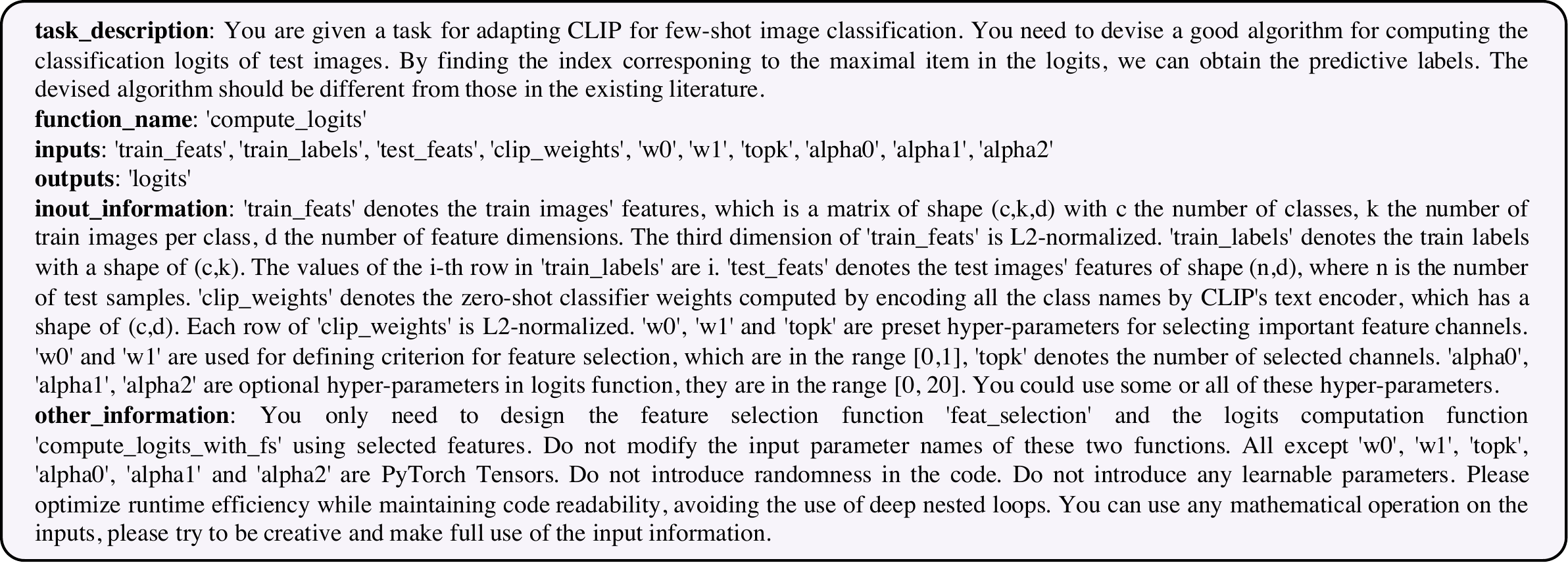}
	\caption{Prompts of crossover and mutation for joint searching. The prompt template is the same to that in Fig.~\ref{fig:prompt}.}
	\label{fig:prompt_joint}
	\vspace{-5pt}
\end{figure*}

The prompt template for joint searching is the same to that in Fig.~\ref{fig:prompt}. However, we slightly change the definition of the keywords in the prompt template, which is given in Fig.~\ref{fig:prompt_joint}. Importantly, in the input parameters, we have also included the hyper-parameters for feature selection. In the `other\_information' part, we instruct LLM to only design the `feat\_selection' and `compute\_logits\_with\_fs' functions.

\begin{table}[thb]
	\tabstyle{7pt}
	\caption{Lookup table for code conversion.}
	\label{tab:code_convert}
	\scriptsize
	\renewcommand{\arraystretch}{1.1}
	\begin{tabular}{lc}
		\toprule
		\rowcolor{white}
		\textbf{Before Conversion} & \textbf{After Conversion} \\
		\midrule
		torch.linalg.pinv & new\_inv = lambda x: torch.linalg.pinv(x.float()).half()\\
		torch.inverse & new\_inv = lambda x: torch.inverse(x.float()).half()\\
		torch.linalg.inv & new\_inv = lambda x: torch.linalg.inv(x.float()).half()\\
		torch.svd & new\_svd = lambda x: torch.svd(x.float()).half()\\
		torch.linalg.eig & new\_eig = lambda x: torch.linalg.eig(x.float()).half()\\
		torch.linalg.eigvals & new\_eigvals = lambda x: torch.linalg.eigvals(x.float()).half()\\
		torch.zeros & new\_zeros = lambda x: torch.zeros(x, dtype=torch.half)\\
		torch.zeros\_like & new\_zeros\_like = lambda x: torch.zeros\_like(x, dtype=torch.half) \\
		torch.ones & new\_ones = lambda x: torch.ones(x, dtype=torch.half)\\
		torch.ones\_like & new\_ones\_like = lambda x: torch.ones\_like(x, dtype=torch.half) \\
		torch.eye & new\_eye = lambda x: torch.eye(x, dtype=torch.half)\\
		F.one\_hot & new\_onehot = lambda x: F.one\_hot(x).half()\\
		\bottomrule
	\end{tabular}
	\vspace{-5pt}
\end{table}
\begin{table}[thb]
	\tabstyle{12pt}
	\caption{Hyper-parameters and searching ranges in the downstream validation stage.}
	\label{tab:hyper_params}
	\scriptsize
	\renewcommand{\arraystretch}{1.1}
	\begin{tabulary}{\linewidth}{JJ}
		\toprule
		\rowcolor{white}
		\textbf{Method} & \textbf{Hyper-parameters and ranges} \\
		\midrule
		\multicolumn{2}{l}{\textbf{Manually-designed Algorithms:}}\\
		Tip-Adapter & $\alpha,\beta\in[10^{-9}, 100]$\\
		APE & $\alpha,\beta,\gamma\in[10^{-9}, 100]$, $topk\in (0,d]$, $w_0$\\
		& and $w_1$ are set according to original paper\\
		GDA & $\alpha \in [10^{-9}, 100]$\\
		\midrule
		\multicolumn{2}{l}{\textbf{Automatically-designed Algorithms:}}\\
		Tip-Adapter/APE/GDA + fs$\rightarrow$logit & $\alpha_0,\alpha_1,\alpha_2\in[10^{-9}, 100]$, $topk\in (0,d]$\\
		& $w_0=w_1=0.5$ \\
		Tip-Adapter/APE/GDA + logit$\rightarrow$fs & $\alpha_0,\alpha_1,\alpha_2\in[10^{-9}, 100]$, $topk\in (0,d]$\\
		& $w_0=w_1=0.5$\\
		Tip-Adapter/APE/GDA + joint & $\alpha_0,\alpha_1,\alpha_2\in[10^{-9}, 100]$, $topk\in (0,d]$\\
		& $w_0=w_1=0.5$\\
		Tip-Adapter/GDA + logit & $\alpha_0,\alpha_1,\alpha_2\in[10^{-9}, 100]$\\
		APE + logit & $\alpha_0,\alpha_1,\alpha_2\in[10^{-9}, 100]$, $topk\in (0,d]$\\
		& $w_0=w_1=0.5$\\
		\bottomrule
	\end{tabulary}
	\vspace{-5pt}
\end{table}

\subsection{More Results}
\subsubsection{Joint Searching}
\begin{figure*}[t]
	\centering
	\includegraphics[width=0.9\linewidth]{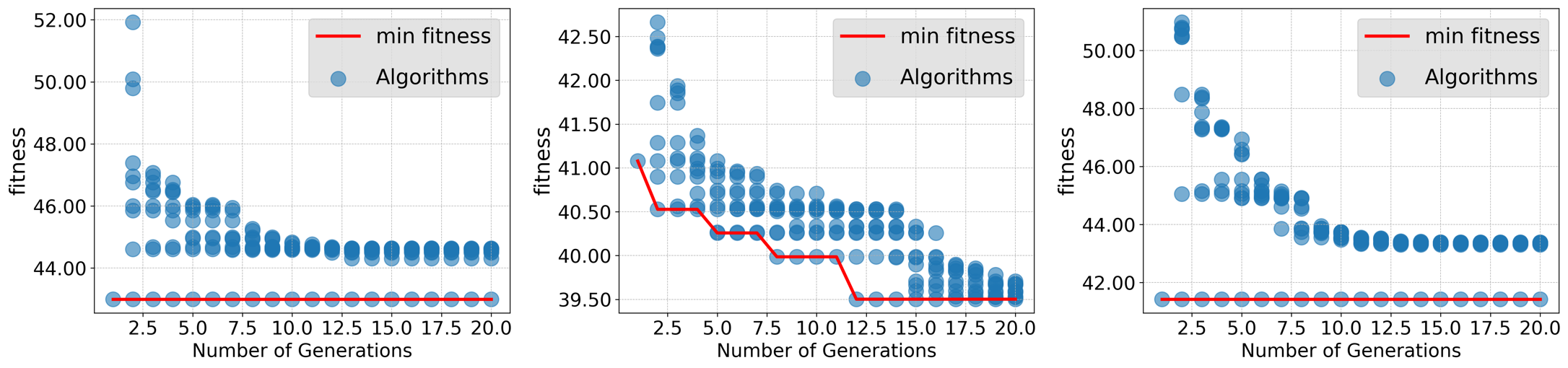}
	\caption{\textbf{Visualization of searching process of joint searching.} The logits computation algorithm is initialized with Tip-Adapter (left), APE (middle) and GDA (right).}
	\label{fig:joint_pops} 
\end{figure*}

The curves of searching process of joint searching can be found in Fig.~\ref{fig:joint_pops}. While the logits computation algorithm is initialized with Tip-Adapter or GDA, we can find that, the minimal objective curve is a horizontal line. This implies that no better algorithm that the initial algorithm can be found in the searching process. We conjecture that the combination of logits computation and feature selection makes it hard to perform searching in the code space.

\subsubsection{Time Cost and Token Cost}
We quantify the computational cost of our search procedure using the DeepSeek API, as detailed in Table~\ref{tab:cost}. The reported token counts encompass both input and output tokens. Our analysis reveals that GDA-based initialization incurs the highest computational overhead in terms of both time and token consumption. Based on the current pricing of \$0.55 per million output tokens, each search iteration costs up to \$0.44 (0.8M tokens $\times $ \$0.55/M). Furthermore, our profiling indicates that the majority of the time expenditure stems from code execution, which involves intensive data processing and iterative operations across datasets, shot configurations, and hyper-parameters. We anticipate that scaling our evaluation infrastructure (currently comprising 2$\times$ RTX 3090 GPUs with 4 parallel services per GPU) could significantly reduce the overall time cost.



\begin{table}[t]
	\tabstyle{2pt}
	\caption{Time and token cost for code searching.}
	\label{tab:cost}
	\renewcommand{\arraystretch}{0.9}
	\begin{tabular}{c c c c c c c c}
		\toprule
		\rowcolor{white}
		\multirow{2}{*}{\textbf{Init}} & \multirow{2}{*}{\textbf{Strategy}}&\multicolumn{2}{c}{\textbf{feature selection}} & \multicolumn{2}{c}{\textbf{logits computation}} & \multicolumn{2}{c}{\textbf{Total}} \\
		\cmidrule(r){3-4}\cmidrule(lr){5-6}\cmidrule(l){7-8}
		&& \textbf{Token/K} & \textbf{Time/h} & \textbf{Token/K }& \textbf{Time/h} & \textbf{Token/K} & \textbf{Time/h}\\
		\midrule
		TIP & fs$\rightarrow$logit & 231 & 0.47 & 402 & 0.76 & 633 & 1.23\\
		TIP & logit$\rightarrow$fs & 365 & 1.91 & 248 & 0.78 & 613 & 2.69\\
		TIP & joint & -- & -- & -- & -- & 1027&3.83\\
		\midrule
		APE & fs$\rightarrow$logit & 256 & 0.98 & 420 & 1.90 & 676 & 2.88\\
		APE & logit$\rightarrow$fs & 500 & 2.24 & 257 & 1.09 & 757 & 3.33\\
		APE & joint & -- & -- & -- & -- & 1044&3.57\\
		\midrule
		GDA & fs$\rightarrow$logit & 279 &1.03& 533 &3.12 & 812 & 4.16\\
		GDA & logit$\rightarrow$fs & 400 &2.13& 298 &1.50 & 698 & 3.64\\
		GDA & joint & -- & -- & -- & -- & 1428&4.40\\
		\bottomrule
	\end{tabular}
	\vspace{-7pt}
\end{table}

\subsubsection{Time Complexity of Searched Code}
We conduct experiments to analyze the time complexity of generated algorithms. The runtime analysis uses the following experimental setup: ResNet-50 architecture, 2,000 training images, test batch size of 256, 100 output classes, and NVIDIA RTX 3090 GPU. The results are presented in Table~\ref{tab:time_comp}, which reveal two key findings: 1) The generated feature selection algorithm demonstrates significant time efficiency improvements, attributable to our explicit prompt design that effectively avoids computationally expensive multi-level loop; 2) While logit computation exhibits a marginal slowdown, this has negligible impact on overall inference time since image feature extraction remains the dominant computational bottleneck. These conclude that the LLM-designed algorithms not only achieve improved accuracy but also maintain computational efficiency.

\begin{table}[t]
	\tabstyle{4pt}
	\caption{Comparison of time complexity of searched code.}
	\label{tab:time_comp}
	\renewcommand{\arraystretch}{1.1}
	\begin{tabulary}{\columnwidth}{cccccccc}
		\toprule
		\rowcolor{white}
		\textbf{stage} & \multicolumn{3}{c}{\textbf{hand-designed}} & \multicolumn{3}{c}{\textbf{Deepseek-generated}}\\
		\cmidrule(lr){2-4}\cmidrule(lr){5-7}
		& \textbf{TIP} & \textbf{APE} & \textbf{GDA} & \textbf{TIP} & \textbf{APE} & \textbf{GDA}\\
		\midrule
		feat extraction (fixed) & 0.0122 & 0.0122 & 0.0122 & 0.0122 & 0.0122 & 0.0122 \\
		feat selection & -- & 0.9111 & -- & 0.0006 & 0.0007 & 0.0005 \\
		logit computation & 0.0004 &	0.0013	& 0.0953 & 0.0012	& 0.0142	& 0.0645 \\
		total time	& 0.0126	& 0.9246	& 0.1075 & 	0.0140	& 0.0271	& 0.0772 \\
		\bottomrule
	\end{tabulary}
	\vspace{-5pt}
\end{table}

\subsubsection{More Per-shot Results}
More per-shot results using different strategies and different combinations of holdout datasets are given in Table~\ref{tab:more_pershot_details}.

\begin{table*}[!thb]
	\tabstyle{14pt}
	\caption{Results of few-shot recognition (ResNet-50). C: CIFAR100, F: FashionMnist, O: ObjectNet, M: UcMerced, B: UCSDBirds}
	\label{tab:more_pershot_details}
	\renewcommand{\arraystretch}{0.9}
	\begin{tabular}{c c c c c c c c c c}
		\toprule
		\rowcolor{white}
		\textbf{Datasets} & \textbf{Initialization} & \textbf{Strategy} & \textbf{1-shot} & \textbf{2-shot}& \textbf{4-shot} & \textbf{8-shot} & \textbf{16-shot} & \textbf{Avg.}\\
		\midrule
		\multirow{4}{*}{C+F+O+M+B}&\multirow{4}{*}{Tip-Adapter}  & fs$\rightarrow$logit & 62.72&	64.93&	67.53	&69.69&	71.17&	67.21\\
		& & logit$\rightarrow$fs & 62.91	&64.76	&67.25&	69.33&	71.00&	67.05\\
		& & joint &62.51	&64.16	&66.52	&69.01&	70.71	&66.58\\
		& & logit & 62.51&	64.77&	67.49&	69.73&	71.54&	67.21\\
		\midrule
		\multirow{4}{*}{C+F+O+M+B}&\multirow{4}{*}{APE}  & fs$\rightarrow$logit & 64.18	&66.23	&69.60&	72.09&	74.11	&69.24\\
		&& logit$\rightarrow$fs & 63.80&	66.07&	69.51&	72.24&	74.00&	69.12\\
		& & joint &64.18	&65.75	&68.45&	71.03&	72.89&	68.46\\
		& & logit & 63.57&	66.14&	69.22&	71.11&	72.71&	68.55\\
		\midrule
		\multirow{4}{*}{C+F+O+M+B}&\multirow{4}{*}{GDA}  & fs$\rightarrow$logit &63.56&	66.51&	70.35&	73.99&	76.43&	70.17\\
		& & logit$\rightarrow$fs &62.99&	66.20&	70.02&	73.59&	76.33&	69.83\\
		&  & joint &62.97&	65.79&	69.85&	72.97&	75.57&	69.43\\
		& & logit & 62.96&	66.06	&70.24&	73.69&	76.23& 69.84\\
		\midrule
		\multirow{2}{*}{C+F+O+M+B}&\multirow{2}{*}{--} & fs$\rightarrow$logit & 62.25&	63.60&	65.79&	67.48&	68.28&	65.48\\
		& & logit$\rightarrow$fs & 61.53&	64.12&	66.56&	68.73&	69.94&	66.18\\
		\midrule
		\midrule
		\multirow{2}{*}{C+F} & \multirow{2}{*}{Tip-Adapter} & fs$\rightarrow$logit & 63.73&	65.14&	68.02&	70.07&	71.58&	67.71\\
		&  & logit$\rightarrow$fs & 63.00&	64.53&	67.53&	69.88&	71.26	&67.24\\
		\midrule
		\multirow{2}{*}{C+F}& \multirow{2}{*}{APE} & fs$\rightarrow$logit &64.01	&65.64	&68.47&	70.84&	72.97&	68.39\\
		& & logit$\rightarrow$fs &63.66	&65.94&	68.74&	71.41&	73.27	&68.60\\
		\midrule
		\multirow{2}{*}{C+F}& \multirow{2}{*}{GDA} & fs$\rightarrow$logit &63.56	&66.51&	70.35	&73.99	&76.43	&70.17\\
		& & logit$\rightarrow$fs &62.99	&66.20&	70.02&	73.59&	76.33&	69.83\\
		\bottomrule
	\end{tabular}
	\vspace{-8pt}
\end{table*}

\subsubsection{Per-dataset Results}
\begin{figure*}[!htbp]
	\centering
	\begin{minipage}{0.23\textwidth}
		\centering
		\includegraphics[width=1.0\textwidth]{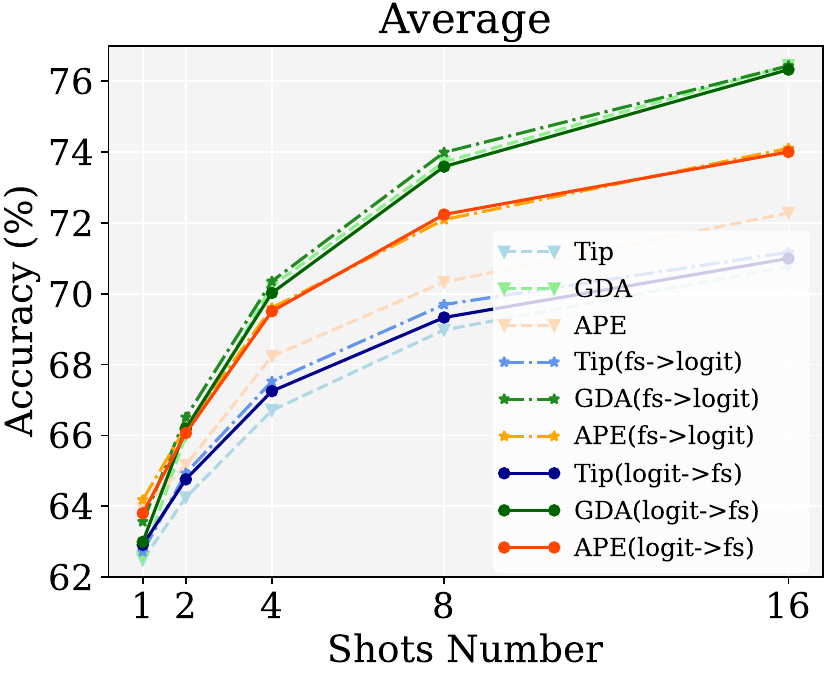}
	\end{minipage}
	\begin{minipage}{0.23\textwidth}
		\centering
		\includegraphics[width=1.0\textwidth]{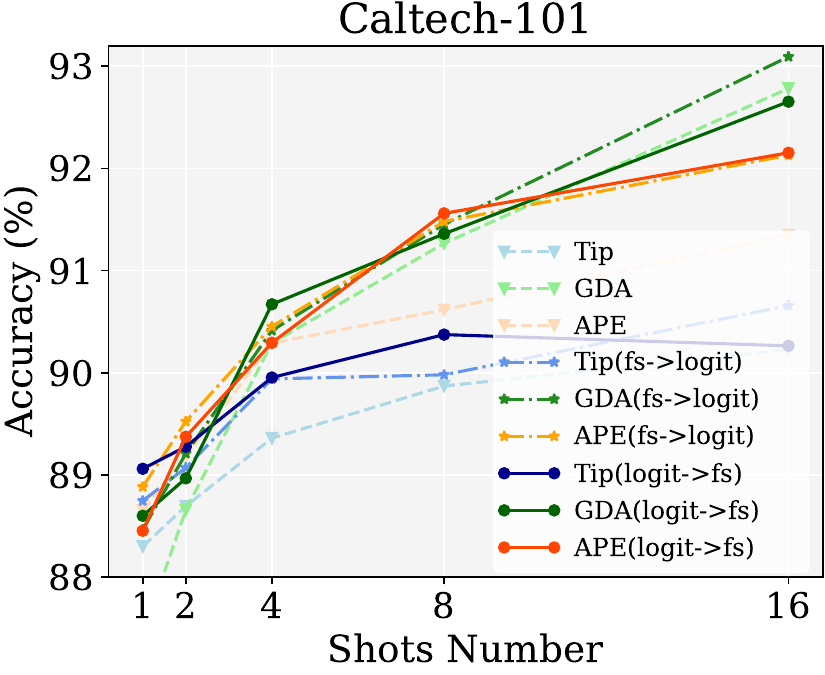}
	\end{minipage}
	\begin{minipage}{0.23\textwidth}
		\centering
		\includegraphics[width=1.0\textwidth]{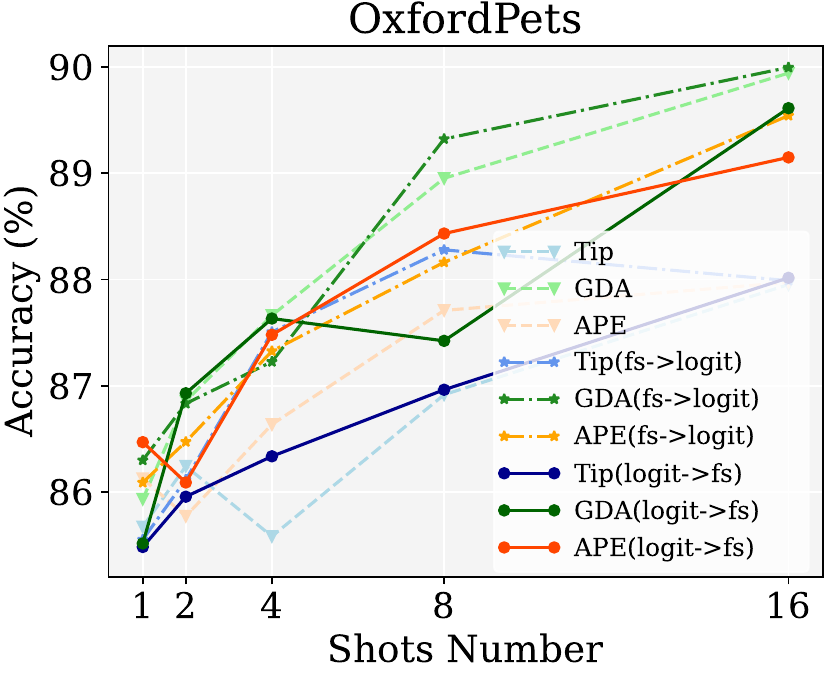}
	\end{minipage}
	\begin{minipage}{0.23\textwidth}
		\centering
		\includegraphics[width=1.0\textwidth]{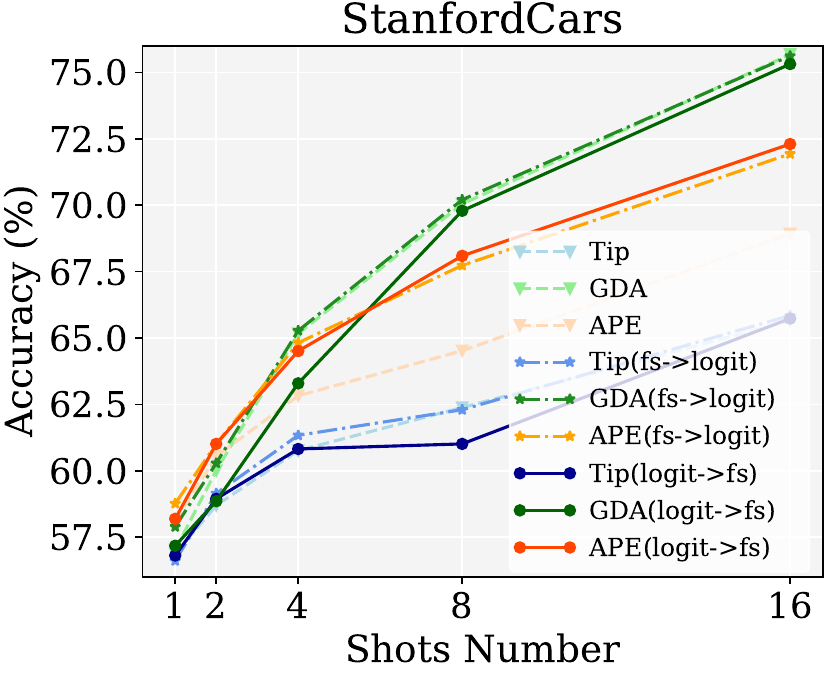}
	\end{minipage}
	\\
	\begin{minipage}{0.23\textwidth}
		\centering
		\includegraphics[width=1.0\textwidth]{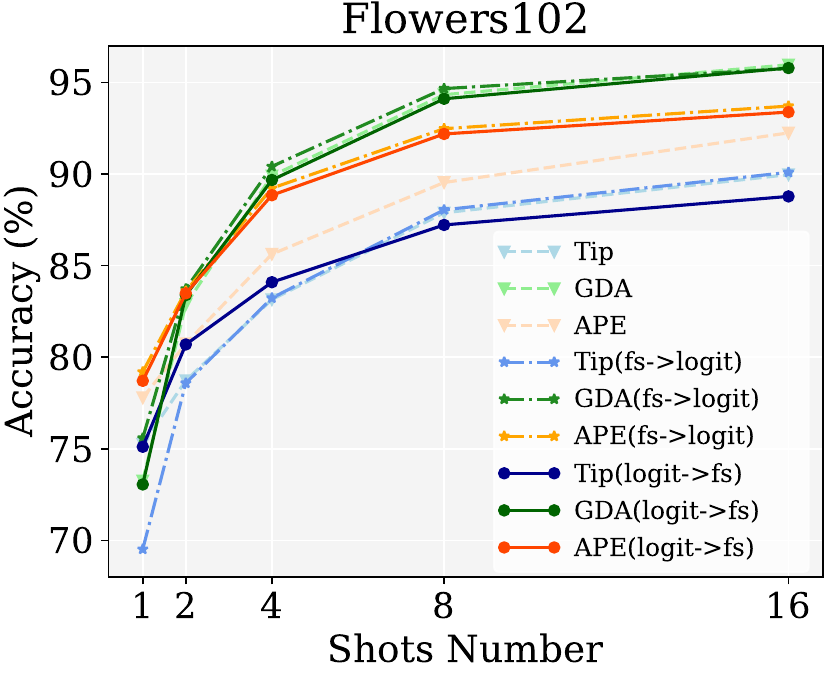}
	\end{minipage}
	\begin{minipage}{0.23\textwidth}
		\centering
		\includegraphics[width=1.0\textwidth]{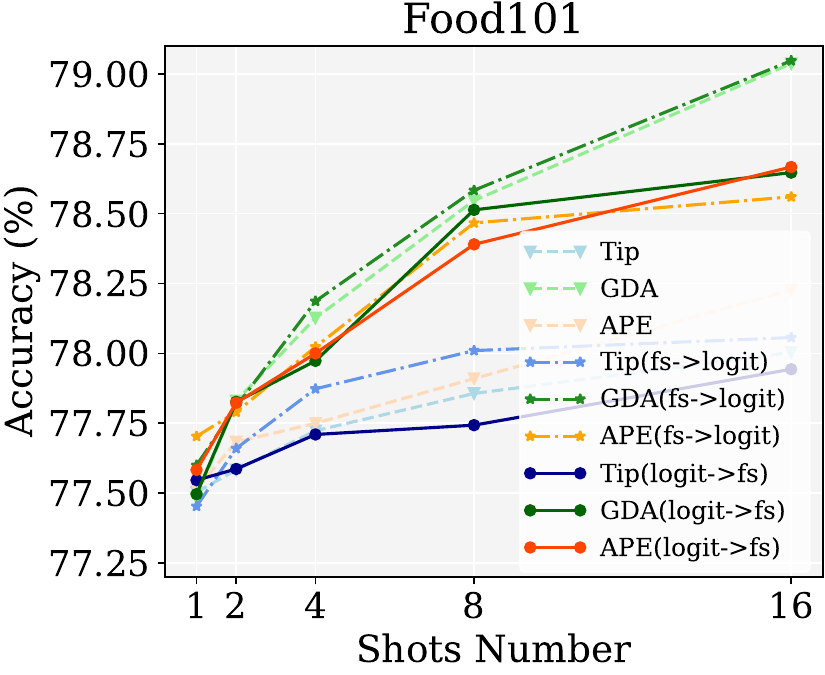}
	\end{minipage}
	\begin{minipage}{0.23\textwidth}
		\centering
		\includegraphics[width=1.0\textwidth]{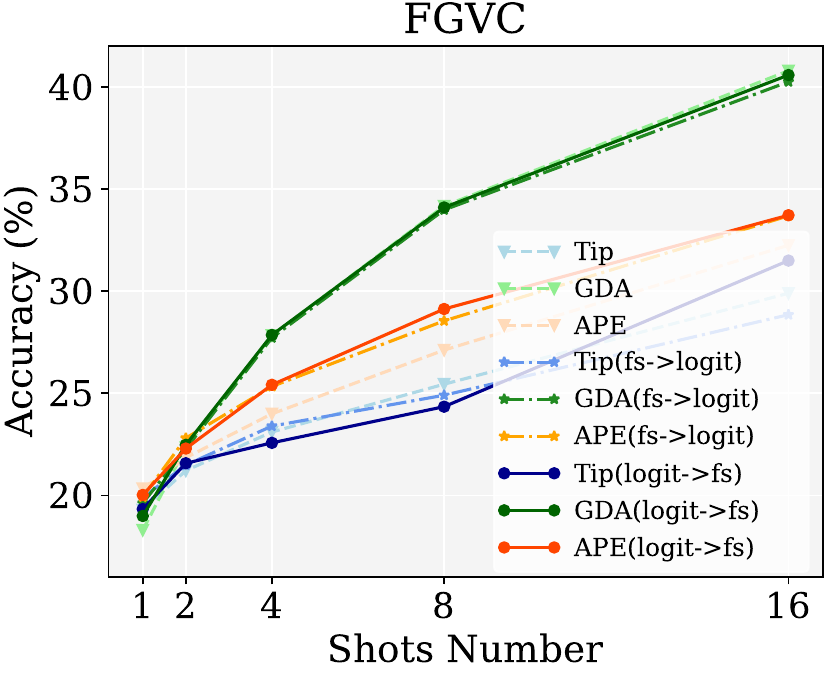}
	\end{minipage}
	\begin{minipage}{0.23\textwidth}
		\centering
		\includegraphics[width=1.0\textwidth]{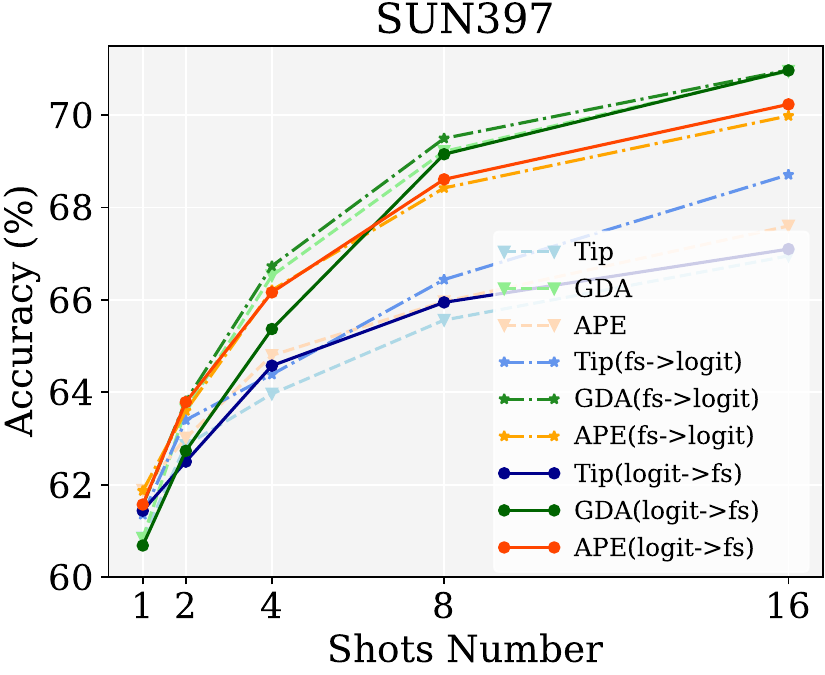}
	\end{minipage}
	\\
	\begin{minipage}{0.23\textwidth}
		\centering
		\includegraphics[width=1.0\textwidth]{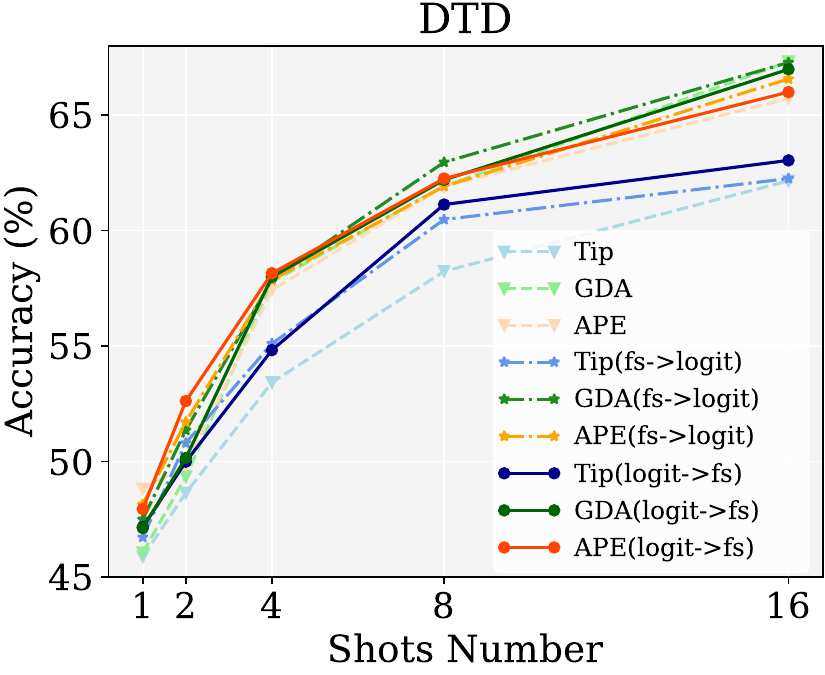}
	\end{minipage}
	\begin{minipage}{0.23\textwidth}
		\centering
		\includegraphics[width=1.0\textwidth]{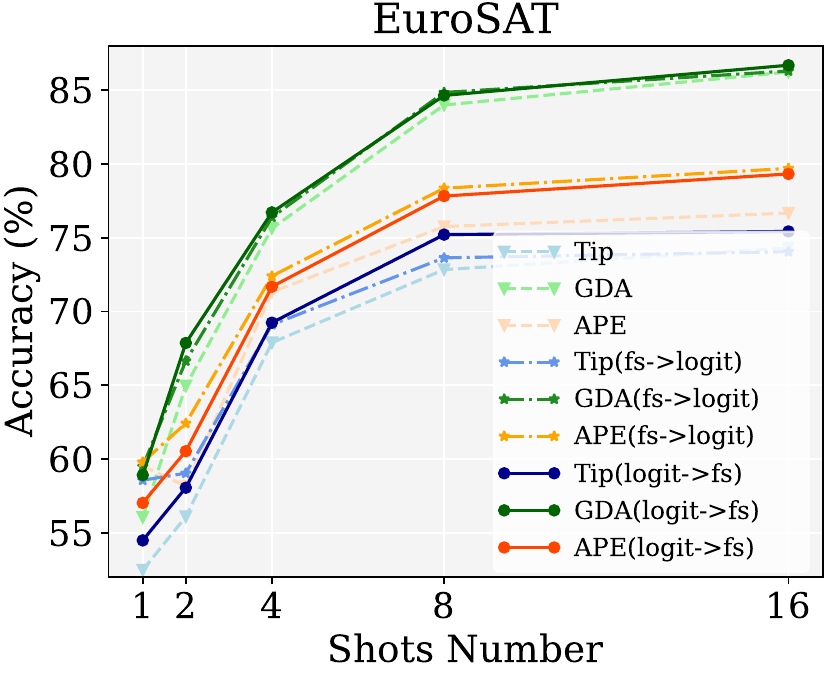}
	\end{minipage}
	\begin{minipage}{0.23\textwidth}
		\centering
		\includegraphics[width=1.0\textwidth]{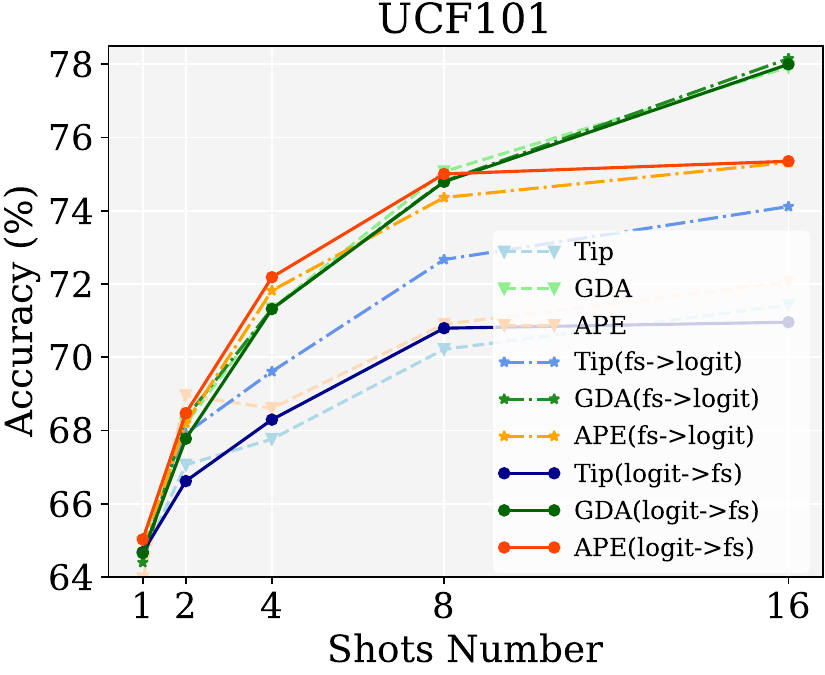}
	\end{minipage}
	\begin{minipage}{0.23\textwidth}
		\centering
		\includegraphics[width=1.0\textwidth]{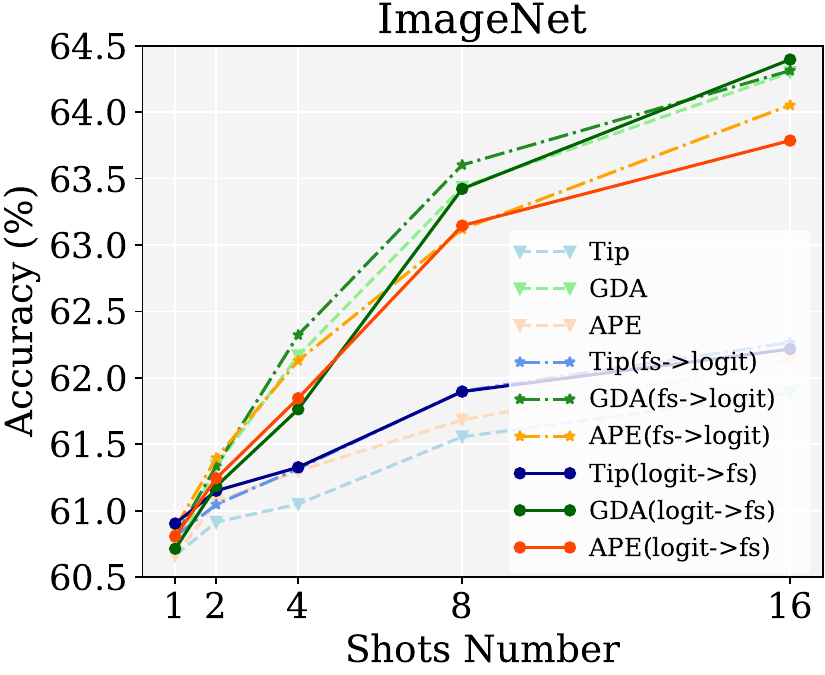}
	\end{minipage}
	\caption{
		{Comparison of per-dataset few-shot classification results.}
	}
	\vspace{-5pt}
	\label{fig:per_data_fs}
\end{figure*}

The per-dataset results of few-shot classification are shown in Fig.~\ref{fig:per_data_fs}. From the figure, we can see that the searched algorithms with APE initialization surpass APE significantly with different shots. However, the searched algorithms with Tip-Adapter or GDA based initialization outperform the baselines slightly. We conjecture that the logits computation algorithm of APE has already used the selected features, which would make it easier for subsequent code searching. By contrast, although we have added the `indices' to the function definition of Tip-Adapter's and GDA's logits computation code, it requires LLM learns to use this input before further improving the usage skills.

\begin{figure*}[!thb]
	\centering
	\includegraphics[width=0.9\linewidth]{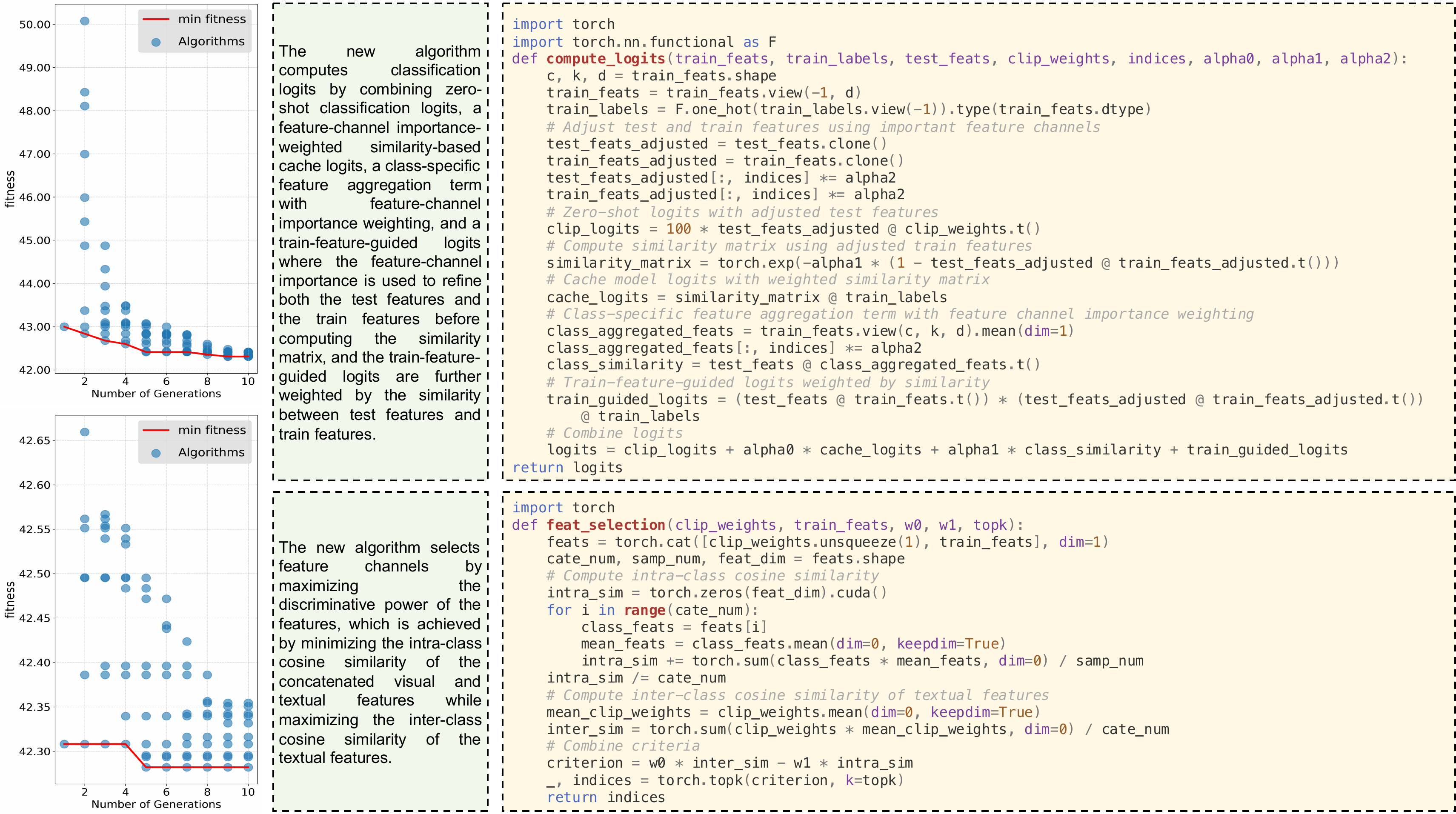}
	\caption{\textbf{Visualization of searching process and results (with Tip-Adapter based initialization).} Top row:  logits computation (first stage), bottom row: feature selection (second stage). From left to right: population at each iteration, thoughts, code.}
	\label{fig:TIP_5di1nt10_logit+fs_vis}
\end{figure*}

\begin{figure*}[thb]
	\centering
	\includegraphics[width=0.85\linewidth]{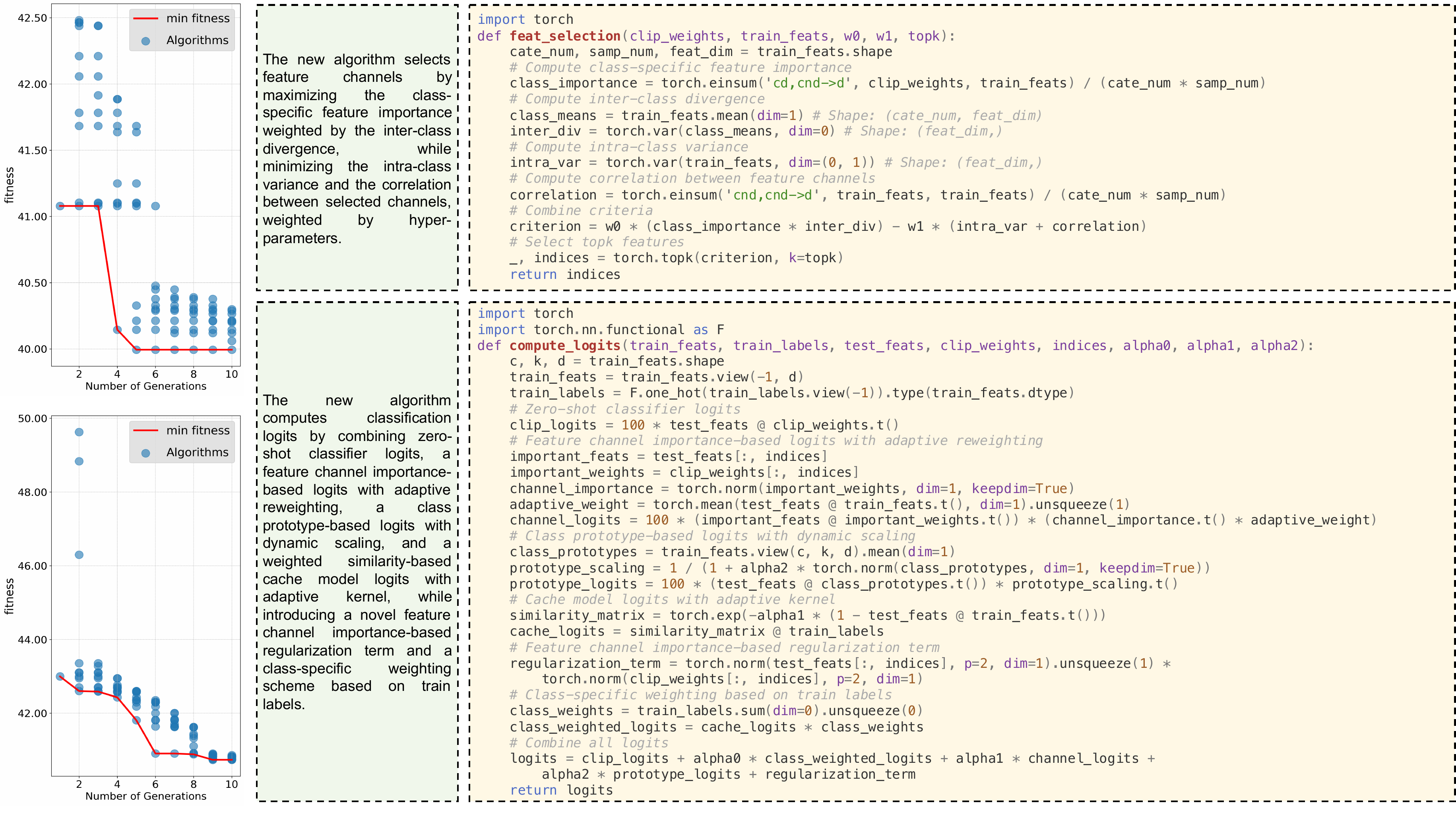}
	\caption{\textbf{Visualization of searching process and results (with Tip-Adapter based initialization).} Top row:  feature selection (first stage), bottom row: logits computation (second stage). From left to right: population at each iteration, thoughts, code.}
	\label{fig:TIP_5di1nt10_fs+logit_vis}
	\vspace{-5pt}
\end{figure*}

\subsubsection{Searched Algorithms}
\begin{figure*}[thb]
	\centering
	\includegraphics[width=0.85\linewidth]{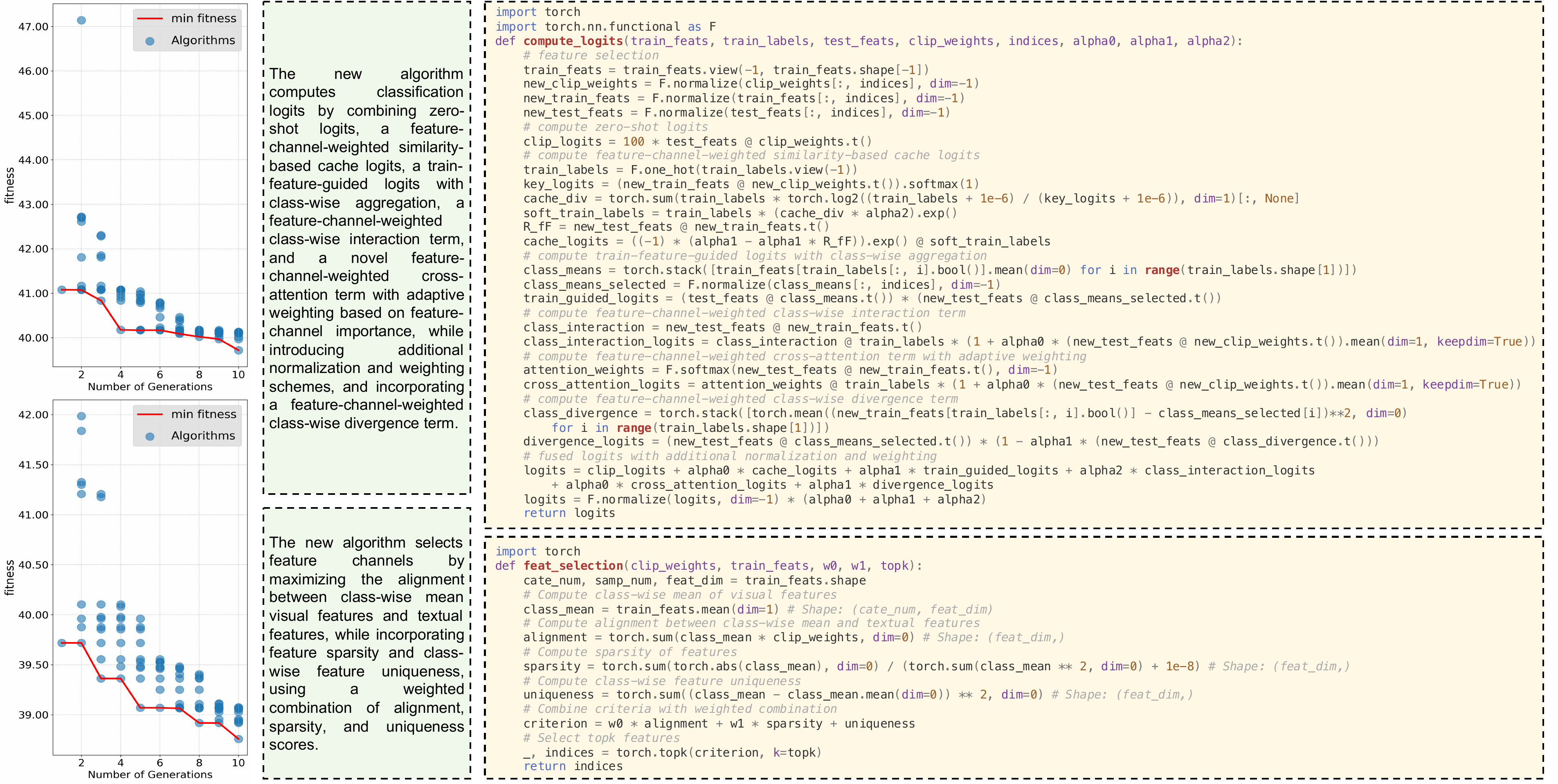}
	\caption{\textbf{Visualization of searching process and results (with APE based initialization).} Top row:  logits computation (first stage), bottom row: feature selection (second stage). From left to right: population at each iteration, thoughts, code.}
	\label{fig:APE_5di1nt10_logit+fs_vis}
	\vspace{-5pt}
\end{figure*}

\begin{figure*}[thb]
	\centering
	\includegraphics[width=0.85\linewidth]{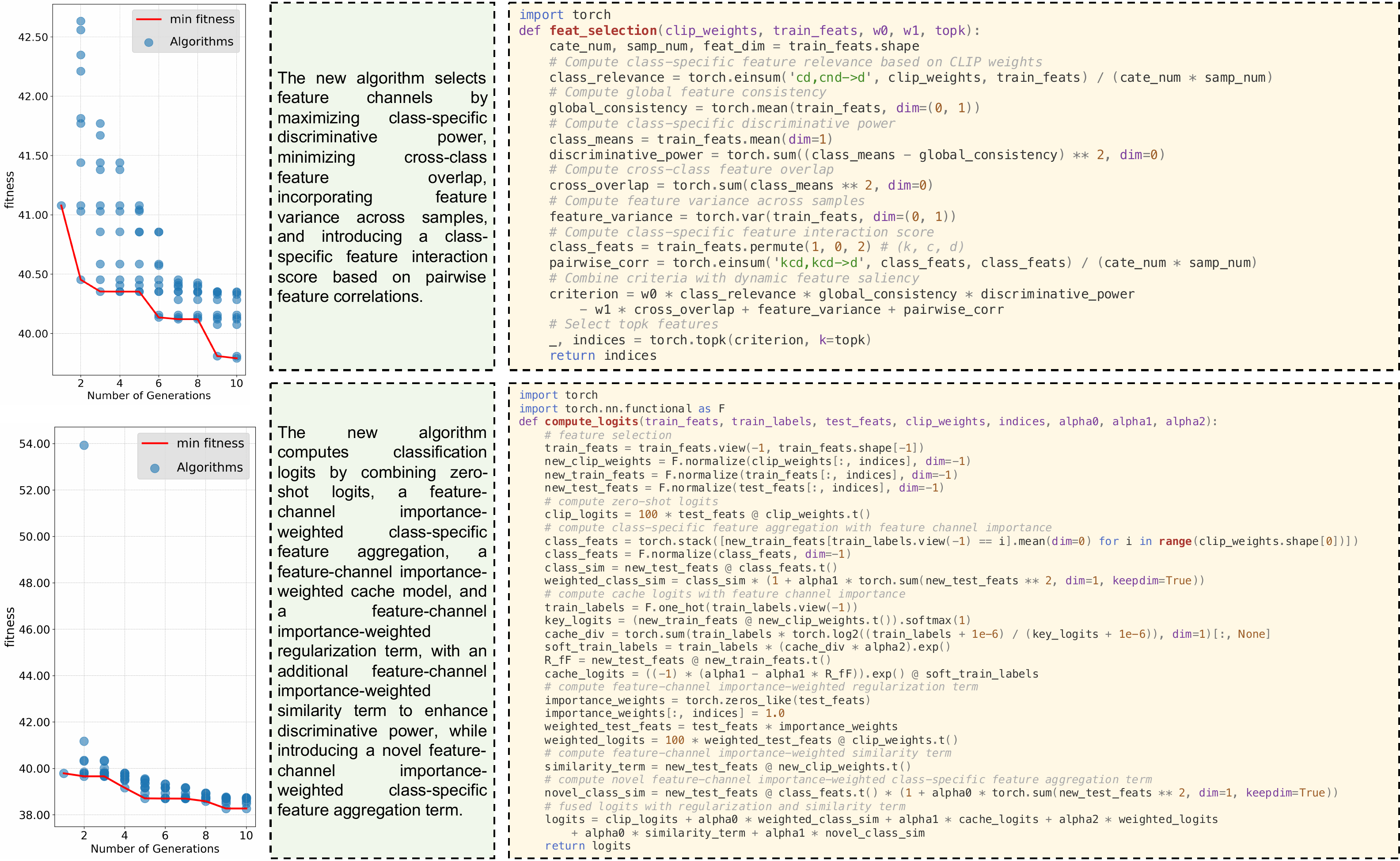}
	\caption{\textbf{Visualization of searching process and results (with APE based initialization).} Top row:  feature selection (first stage), bottom row: logits computation (second stage). From left to right: population at each iteration, thoughts, code.}
	\label{fig:APE_5di1nt10_fs+logit_vis}
	\vspace{-5pt}
\end{figure*}

\begin{figure*}[t]
	\centering
	\includegraphics[width=0.85\linewidth]{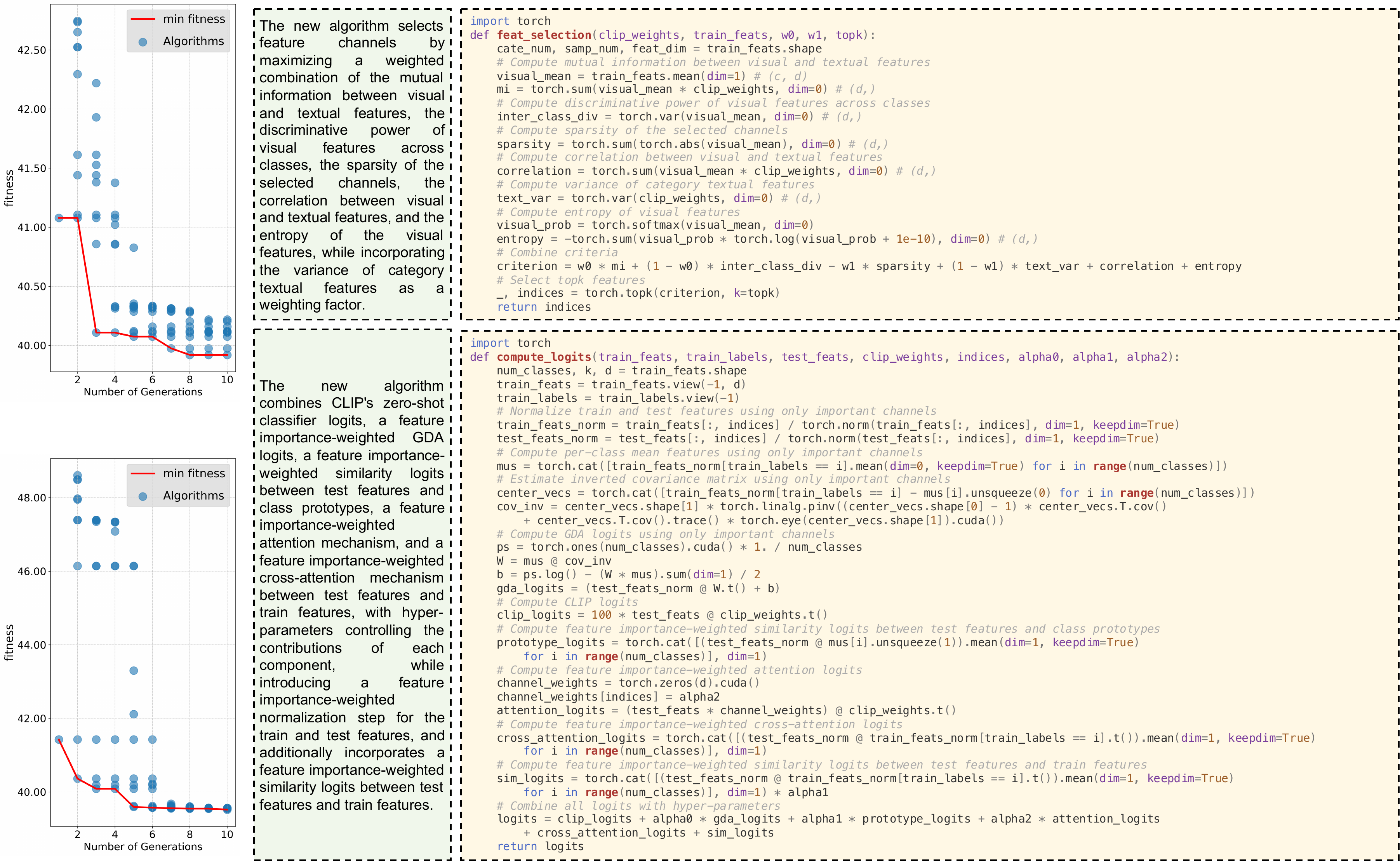}
	\caption{\textbf{Visualization of searching process and results (with GDA based initialization).} Top row:  feature selection (first stage), bottom row: logits computation (second stage). From left to right: population at each iteration, thoughts, code.}
	\label{fig:GDA_5di1nt10_fs+logit_vis}
	\vspace{-5pt}
\end{figure*}

Here, we visualize more results of searched algorithms with different initialization in Fig.~\ref{fig:TIP_5di1nt10_logit+fs_vis}, Fig.~\ref{fig:TIP_5di1nt10_fs+logit_vis}, Fig.~\ref{fig:APE_5di1nt10_logit+fs_vis}, Fig.~\ref{fig:APE_5di1nt10_fs+logit_vis} and Fig.~\ref{fig:GDA_5di1nt10_fs+logit_vis}.

In Fig.~\ref{fig:TIP_5di1nt10_logit+fs_vis}, the searched algorithms under the strategy `logit$\rightarrow$fs' using Tip-Adapter's logits computation algorithm and APE's feature selection algorithm as initialization are shown. The searched logits computation algorithm has four terms: `clip\_logits', `cache\_logits', `class\_similarity' and `train\_guided\_logits'. Instead of directly selecting the important features by the indices, this algorithm innovatively introduces a hyper-parameter as a multiplier for important features. The `class\_similarity' computes logits using class aggregated train features. Besides, the `train\_guided\_logits' term essentially utilizes the second-order form of the similarity matrix to enhance non-linearity. The searched feature selection algorithm has confused inter-class and intra-class similarity. Meantime, the code fails to correctly compute the intra-class and inter-class similarity. Therefore, to some extent, the code generation capability of current LLMs still has flaws.

In Fig.~\ref{fig:TIP_5di1nt10_fs+logit_vis}, the searched algorithms under the strategy `fs$\rightarrow$logits' using Tip-Adapter's logits computation algorithm and APE's feature selection algorithm as initialization are shown. The searched feature selection algorithm innovatively introduces the concept of class importance and it has been correctly implemented by code. It essentially reflects the alignment between textual and visual features, which is of good rationality. The `inter\_div' and `intra\_var' terms are relatively conventional designs. The algorithm also comes up with a correlation term, however, the corresponding code implementation is wrong. As for the logits computation algorithm, there are five terms in the designed algorithm. The `class\_weighted\_logits' is the product of `cache\_logits' and `class\_weights'. However, the introduction of the class weights is useless as we have told LLM that each class has the same number of train samples.

In Fig.~\ref{fig:APE_5di1nt10_logit+fs_vis}, the searched algorithms under the strategy `logit$\rightarrow$fs' using APE's algorithm as initialization are shown. The searched logits computation algorithm has combined six terms. In the cache logits, the selected feature channels are adopted. The `train\_guied\_logits' is a productive composite of two logits: logits using class mean as projection matrix and the same term with only selected features used. This form could be useful for improving the nonlinearity of the classifier. The `class\_interaction\_logits' term is slightly complex, which is a composite of two kinds of logits: a cache-like logits without exp function, a `clip\_logits' like logits with only selected features used. The `cross\_attention\_logits' has similar form to `class\_interaction\_logits', but introduces the softmax function to the train-test similarity matrix. The `divergence\_logits' considers the information of per-class feature divergence. Besides, the logits also undergoes a transform of normalization and re-scaling. However, this line of code is redundant as it does not change the index of maximal items. As for the feature selection algorithm, the automatically-designed algorithm has a concise form, which avoids the nested loops. The designed criterion has three terms: `alignment', `sparsity' and `uniqueness'. The alignment term requires the selected features should be aligned well between visual and textual modalities. The `uniqueness' term is also a variance term, which selects the features of high variance. The `sparsity' term may have little effect as the features may not be sparse. Meantime, the reversal of the sign of the sparsity term does not achieve the original intention of the algorithm.

In Fig.~\ref{fig:APE_5di1nt10_fs+logit_vis}, the searched algorithms under the strategy `fs$\rightarrow$logit' using APE's algorithm as initialization are shown. In the searched feature selection algorithm, the first term emphasizes the correlation between visual and textual features, as well the variance of feature channels. The second, third and fourth terms actually represent the feature variance. As such, there are discrepancies between the code and the comments. As for the logits computation algorithm, modulating the logits by the test features' norm could be a relatively novel design, which are reflected in the `weighted\_class\_sim' and `novel\_class\_sim' terms.

\end{document}